\documentclass[fleqn,10pt]{wlscirep}

\usepackage{amsmath}
\usepackage{nicefrac}
\usepackage[utf8]{inputenc}
\usepackage[T1]{fontenc}
\usepackage[normalem]{ulem}
\usepackage[version=4]{mhchem}
\usepackage{bibunits}
\usepackage{caption}

\newcommand{\dirtyspacingtrick}{\textcolor{white}{XXXX X\hspace{1pt}}}

\title{The unrealized potential of agroforestry for an emissions-intensive agricultural commodity}

\author[1]{Alexander Becker}
\author[2]{Jan D. Wegner}
\author[3]{Evans Dawoe}
\author[1]{Konrad Schindler}
\author[4]{William J. Thompson}
\author[5]{Christian Bunn}
\author[6]{Rachael D. Garrett}
\author[5]{Fabio Castro-Llanos}
\author[7,*]{Simon P. Hart}
\author[7,*,$@$]{Wilma J. Blaser-Hart}
\affil[1]{~Photogrammetry and Remote Sensing, ETH Zurich, Switzerland}
\affil[2]{~EcoVision Lab, Dept.\ of Mathematical Modeling and Machine Learning, University of Zurich, Switzerland}
\affil[3]{~Dept.\ of Agroforestry, Faculty of Renewable Natural Resources, Kwame Nkrumah University of Science and \dirtyspacingtrick Technology, Kumasi, Ghana}
\affil[4]{~Nature-based Solutions Initiative, Department of Biology and Smith School of Enterprise and the Environment, University of Oxford}
\affil[5]{~International Center for Tropical Agriculture, Cali, Colombia}
\affil[6]{~Conservation and Development Lab, Department of Geography and Conservation Research Institute, University of Cambridge, Cambridge, UK}
\affil[7]{~School of the Environment, and Centre for Biodiversity and Conservation Science (CBCS), The University of Queensland, St Lucia, QLD 4072, Australia\vspace{.5em}}

\affil[$\!\!@\;$]{w.hart@uq.edu.au}
\affil[*]{These authors contributed equally}

\begin{document}

\flushbottom
\maketitle
\thispagestyle{empty}
\section*{Abstract}
Reconciling agricultural production with climate-change mitigation is a formidable sustainability problem. Retaining trees in agricultural systems is one proposed solution, but the magnitude of the current and future-potential benefit that trees contribute to climate-change mitigation remains uncertain. Here, we help to resolve these issues across a West African region that produces $\sim$60\% of the world's cocoa, a crop contributing one of the highest carbon footprints of all foods. Using machine learning, we mapped shade-tree cover and carbon stocks across the region and found that existing average cover is low ($\sim$13\%) and poorly aligned with climate threats. Yet, increasing shade-tree cover to a minimum of 30\% could sequester an additional 307~million~tonnes of \ce{CO2}e, enough to offset $\sim$167\% of contemporary cocoa-related emissions in Ghana and Côte d’Ivoire—without reducing production. Our approach is transferable to other shade-grown crops and aligns with emerging carbon market and sustainability reporting frameworks.

\section*{Introduction}
Integrating trees into agricultural systems has been widely recognized as one of the most effective nature-based solutions, offering co-benefits for climate-change mitigation through carbon storage, climate-change adaptation via microclimate regulation, and support for biodiversity and ecosystem services \cite{blaser18,griscom2017,bennett2022,terasaki2023}. Nevertheless, the magnitude of the current benefit that trees bring for climate and conservation, as well as their potential future benefit, has been controversial and remains uncertain \cite{phalan2011,balmford2018,terasaki2023,sills2019,veldman2019,bastin2019}. This uncertainty persists even though agroforestry is included in the Nationally Determined Contributions (NDCs) of around 40\% of non-Annex I countries under the Paris Agreement, as a strategy to reduce greenhouse gas emissions\cite{rosenstock2019}, and despite numerous large-scale and industry-specific initiatives aiming to increase the number and cover of trees both locally and globally \cite{brancalion2020,holl2020}. There is, therefore, an urgent need to develop cost-effective, accurate, and repeatable methods for generating high-resolution assessments of the status of trees in agricultural systems \cite{terasaki2023}. Accounting requirements for carbon markets, as well as emerging sustainability reporting requirements for corporations (e.g. \href{https://eur-lex.europa.eu/legal-content/EN/TXT/PDF/?uri=OJ:L_202302772}{EU}, \href{https://www.gov.uk/guidance/uk-sustainability-reporting-standards}{UK}, and \href{https://aasb.gov.au/news/exposure-draft-ed-sr1-australian-sustainability-reporting-standards-disclosure-of-climate-related-financial-information/}{Australia}), only make this need more urgent.

Here we generate high-resolution, spatially-explicit maps of shade-tree cover and biomass (carbon storage) across 7.04 million hectares of West African agricultural systems currently responsible for producing $\approx$60\% of the global supply of cocoa \cite{FAO22}. Quantifying the current and future status of trees in agricultural systems is particularly important given their sheer scale in the historically carbon-dense and biodiverse tropics and subtropics, and the unique potential of agroforestry for allowing climate-smart, sustainable agricultural intensification for a wide range of crops \cite{blaser18,kalischek23natfood}. Cocoa production systems themselves are an important test case. On the one hand, cocoa production has been responsible for widespread historical and contemporary deforestation, including in protected areas \cite{kalischek23natfood,Renier_2023,barima2016}, and cocoa contributes to one of the most emissions-intensive footprints of all foods (\href{https://ourworldindata.org/food-choice-vs-eating-local}{Ritchie (2020)}\cite{poore2018}. On the other hand, studies done on small scales demonstrate that cocoa can be produced together with moderate densities of shade trees (up to {$\approx$30-50}\% cover), with substantial benefits for carbon sequestration and biodiversity, and crucially, with little or no loss of yield \cite{blaser18,andres2018,clough2011,ramirez-argueta2022,bisseleua2009,schroth2016c}. In combination, these impacts and features of cocoa production systems have led to a number of ambitious industry commitments based around the planting of millions of shade-trees in agroforests to improve the sustainability of the sector \cite{carodenuto2021}(e.g. \href{https://worldcocoafoundation.org/programmes-and-initiatives/cocoa-and-forests-initiative}{Cocoa \& Forests Initiative [CFI]}). Yet, while the nature of cocoa production is central to the socio-economic, climate, and biodiversity futures of this entire region, the current status and future potential of these systems is unknown. 

To solve this problem we use machine learning to generate estimates of the cover and biomass of trees in cocoa agroforestry systems across Ghana and Côte d’Ivoire in West Africa (Methods). We generate maps of shade-tree cover and biomass using information in publicly-available satellite imagery, recent global maps of tree-canopy height \cite{lang2023high}, and an extensive ground-truth field campaign conducted over two years across the two countries. Our  approach generates high-resolution, spatially-explicit estimates, tailored to a single crop growing across two countries and five different agro-climatic zones, and includes estimates of mapping uncertainty. This information enables an assessment of the status of agroforestry adoption in cocoa production systems. Then, by generating an empirical relationship between shade-tree cover and biomass, we also generate estimates of the current status and future potential of carbon sequestration in cocoa production systems across the region (Methods). Our results demonstrate the enormous and currently unrealized potential for climate-smart, sustainable intensification for this important global commodity. Moreover, beyond consequences for cocoa production, we have developed methods and a workflow to assess the status of trees in agricultural systems that is repeatable and generally applicable across the tropics and subtropics, including for other globally significant commodities that can be productively and sustainably grown in agroforests (\textit{e.g.} coffee). 

\section*{Results and discussion}

\subsection*{Shade-tree cover is low, and misaligned with climate threats}
Despite the widely acknowledged benefits of agroforestry for cocoa production and long-standing industry commitments to increase shade-tree cover in these systems, we found that cocoa production is overwhelmingly dominated by full-sun monocultures and low-shade agroforestry (Fig.~\ref{fig:shade_map}; see Supplementary Methods for model performance metrics). Average shade-tree cover across both countries was $13.2\pm 9.2\%$ (mean$\pm$standard deviation; Ghana: $13.8\pm 9.2\%$; Côte d'Ivoire: $12.8\pm 9.2\%$). More than 80\% of cocoa production is implemented on farms with less than 20\% shade, and only 5.6\% (396,422 ha) of cocoa is grown under shade levels at or above 30\% cover (Fig.~\ref{fig:shade_map}). Farms with the highest levels of shade-tree cover (above 30\%) tend to occur in small, isolated patches, particularly in the east of Ghana and the north-west of Côte d’Ivoire (Fig.~\ref{fig:shade_map}), and do not tend to overlap with recently deforested areas \cite{kalischek23natfood}. In general, observed cover levels do not compare favourably with common recommendations of between 20\% and 40\% cover\cite{blaser18}, and even up to 50\% cover, for cocoa production systems in this region (\href{https://voicenetwork.cc/wp-content/uploads/2020/07/200706-Cocoa-Barometer-Agroforestry-Consultation-Paper.pdf}{Sanial et al. (2020)}). 

\begin{figure}[h]
    \centering
    \includegraphics[width=\textwidth,trim={0px 0px 0px 0px},clip]{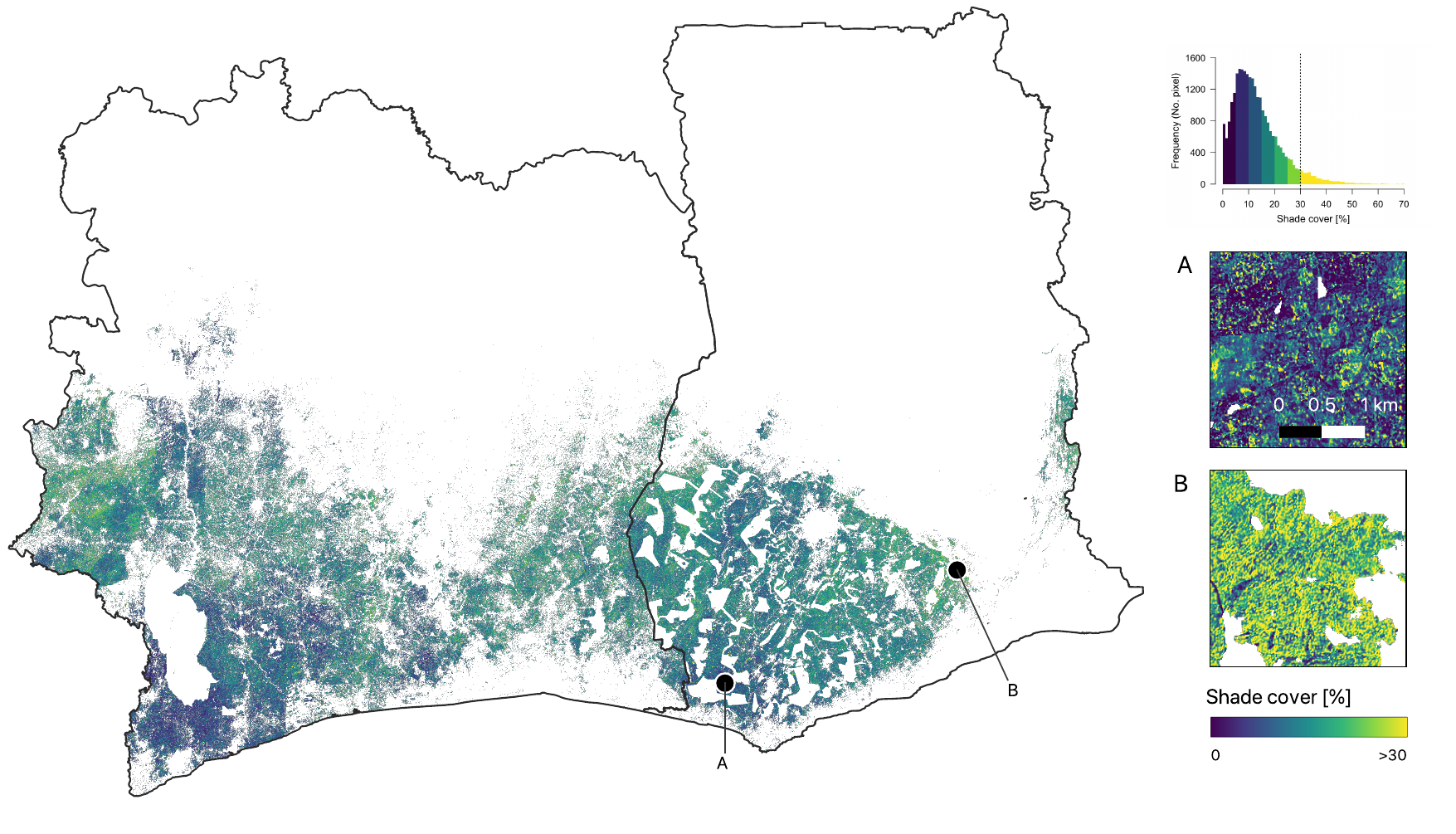}
    \caption{\textbf{Map of shade-tree cover across cocoa-growing areas in Côte d'Ivoire and Ghana for the year 2022}. The shade-cover estimates were derived from Sentinel-2 imagery at a ground sampling distance of $10 \times 10$\,m. For visualization, the map presented here was resampled to a coarser grid of $100 \times 100$\,m. Predictions were restricted to areas mapped as cocoa-growing regions using the cocoa mask from Kalischek et al. (2023)\cite{kalischek23natfood}. The histogram illustrates the distribution of shade-tree cover based on the original $10 \times 10$\,m resolution data ($n = 679,593,915$ individual pixels) across all cocoa-growing areas. Location \textbf{A} depicts an area of cocoa monoculture in the Western Region, while Location \textbf{B} depicts an area of cocoa agroforestry in the Eastern Region of Ghana based on the original $10 \times 10$\,m resolution map. Base map boundaries were derived from the Global Administrative Areas database (GADM v4.1, \href{https://gadm.org/}{https://gadm.org}), used under academic license.}
    \label{fig:shade_map}
\end{figure}

Surprisingly, and counter to expectations and previous research\cite{knudsen2015,thompson2022,ruf2015}, levels of shade-tree cover are not only low, but also do not vary substantially across larger spatial scales, including between administrative regions and agro-climatic zones. For example, average cover levels across agro-climatic zones range between 11.5-15\% (Extended Data Fig.~\ref{fig:aez}), representing a maximum average difference of only 3.5\% between the zones. The highest average cover occurred in the moderate-very dry zone (Type 1) in the northern cocoa growing area, with cover levels gradually declining through the dry and wet zones, reaching the lowest average cover in the very hot-very wet coastal zone (Type 5). Specifically, shade cover decreased from 15.0\% in the moderate-very dry zone to 13.6\%, 12.4\%, and 11.5\% in progressively wetter zones, while the cooler, wet zone (Type 4) had an average cover of 14.1\%. Similar small differences were found between administrative regions with average cover levels ranging between 9 and 17\% (Extended Data Fig. \ref{fig:regions_shade}). The lack of substantial differences across larger spatial scales is surprising for two reasons. First, differences in the timing of the introduction of cocoa production among regions might have been expected to result in differences in shade cover among regions\cite{knudsen2015,thompson2022,ruf2015}. By contrast, the relative uniformity in cover that we document suggests that any such systematic differences across large scales have disappeared. This could be because of the widespread adoption of cocoa monocultures across all regions, and/or because of more spatially-targeted efforts to increase shade levels in areas most dominated by monocultures. Both processes could have reduced historical differences in cover between regions. Regardless of the mechanism, our results emphasize that low shade-tree cover is a pervasive issue that is not limited to recently established farms but affects cocoa-growing landscapes across both countries. Second, higher levels of average cover might have been expected to occur in areas that experience hotter climates as long as they are not too dry\cite{Abdulai2018}, which are also the areas where the persistence of the crop is most vulnerable to climate change (Extended Data Fig.~\ref{fig:aez})\cite{schroth2016b,schroth2017}. Higher average temperatures are predicted to limit cocoa production in the future, emphasizing the importance of the ability of shade-trees to buffer the effects of climate on the crop \cite{schroth2016b,blaser18,niether2020}. The absence of higher shade-tree cover in these more vulnerable areas suggests that farmers are not currently capitalizing on the climate buffering benefits of shade trees, nor are they maintaining higher shade tree densities as a source of alternative income in these areas\cite{schroth2017,schroth2016b,bunn19}. 

Despite the low levels of shade-tree cover that we demonstrate, some of our estimates compare favourably with those used in official reporting. The \href{https://ghana-national-landuse.knust.ourecosystem.com/interface/}{national land use map for Ghana} produced by the Ghana Forestry Commission provides the only other estimates of shade-tree cover across this region, but this map resolves land use only coarsely and distinguishes only between unshaded (< 15\% cover) and shaded (>15\% cover) cocoa systems in line with Ghana’s REDD+ requirements (Ashiagbor, personal communication\cite{epa2007,ashiagbor2020}). Based on this map and classification, 13.6\% of the cocoa growing area was considered "shaded" in 2021. By contrast, our map suggests that 37.2\% of the cocoa-growing region in Ghana meets the REDD+ requirement of 15\% cover, a nearly 3-fold increase over the existing estimate. A discrepancy of this magnitude is not surprising given our technical advance, and while our results demonstrate that shade-tree cover is generally low, this large positive discrepancy may nevertheless have important policy and economic implications.

\subsection*{Cocoa systems have major climate-change mitigation potential}
One of the major benefits of trees in agricultural systems is climate-change mitigation through carbon sequestration \cite{chapman2020,zomer2016,nair2012,terasaki2023,griscom2017}. Past research on small scales has shown that cocoa agroforests can sequester substantial amounts of carbon in the aboveground biomass of trees, and that these benefits increase with increasing shade-tree cover \cite{nair2012,blaser18,niether2020,asigbaase2021}. Here we extend these small-scale studies to estimate carbon storage across the sector in this region (Methods). We find that cocoa production systems across Côte d'Ivoire and Ghana currently store an estimated  $117.7\pm 1.2 \times 10$\textsuperscript{-5} million tonnes of carbon (estimate $ \pm  $ 95\% CI) in above ground biomass. The majority of this carbon is distributed across a large area with relatively low carbon density, averaging $13.7\pm 6.9$ tonnes C ha\textsuperscript{-1} (estimate $ \pm $ SD; Fig.~\ref{fig:agbd_map}, Extended Data Fig.~\ref{fig:landuse_carbon_stocks}). 

Given that levels of shade-tree cover were lower than expected, we estimated the carbon sequestration potential in this system under different agroforestry adoption scenarios aligned with industry commitments (Methods, \href{https://voicenetwork.cc/wp-content/uploads/2020/07/200706-Cocoa-Barometer-Agroforestry-Consultation-Paper.pdf}{Sanial et al. (2020)}).  Increasing shade-tree cover across all farms from current levels to the minimum required for REDD+ compliance in Ghana (15\% cover) would allow for an additional 17 million tonnes of carbon storage, equivalent to 64 million tonnes \ce{CO2}e (Fig.~\ref{fig:C_potential}). If shade-tree cover was increased to a minimum of 30\%, which is in line with emerging recommendations \cite{blaser18,bennett2022,andres2018,ramirez-argueta2022,clough2011} and certification requirements by the Rainforest Alliance (for example), carbon storage would rise by an additional 84 million tonnes of carbon, or 307 million tonnes \ce{CO2}e (Fig.~\ref{fig:C_potential}). Crucially, several studies now demonstrate that shade-tree cover levels of up to $\approx$30-50\% do not necessarily limit cocoa production, allowing these benefits to be achieved without threatening production and livelihoods \cite{blaser18,andres2018,ramirez-argueta2022,bisseleua2009,asare2018,schroth2016c}. 

Assuming linear rates of implementation and carbon sequestration from 2025, the additional carbon storage capacity would allow $\approx$10.2 million tonnes \ce{CO2}e to be sequestered annually for $\approx$30 years, which is $\approx$9.1\% of the current total annual emissions of both countries combined (Methods). For the cocoa industry, these values would allow the industry to annually counterbalance $\approx$167\% of annual emissions (\ce{CO2}e) generated in the production of the crop. Reflecting the high rates of deforestation associated with cocoa production \cite{kalischek23natfood}, this number is much smaller---$\approx$15\%---when accounting for the contribution to the industry's annual emissions of land-use change, particularly deforestation. Nevertheless, these amounts are noteworthy explicitly because cocoa currently contributes to one of the most emissions-intensive footprints of all foods (\href{https://ourworldindata.org/food-choice-vs-eating-local}{Ritchie (2020)})\cite{poore2018}. Moreover, these estimates do not include the substantial additional carbon that would be stored as tree biomass below ground, which can be as high as $\approx$20-35\% of total carbon in tropical systems \cite{ma2021,jackson1996}. These estimates underscore the enormous potential of agroforestry for climate-change mitigation for a single, agricultural sector across the two major producing countries for this crop.

\subsection*{Forests remain key to carbon storage in cocoa landscapes}
\begin{figure}[ht]
    \centering
    \includegraphics[width=0.93\textwidth,trim={20px 30px 10px 30px},clip]{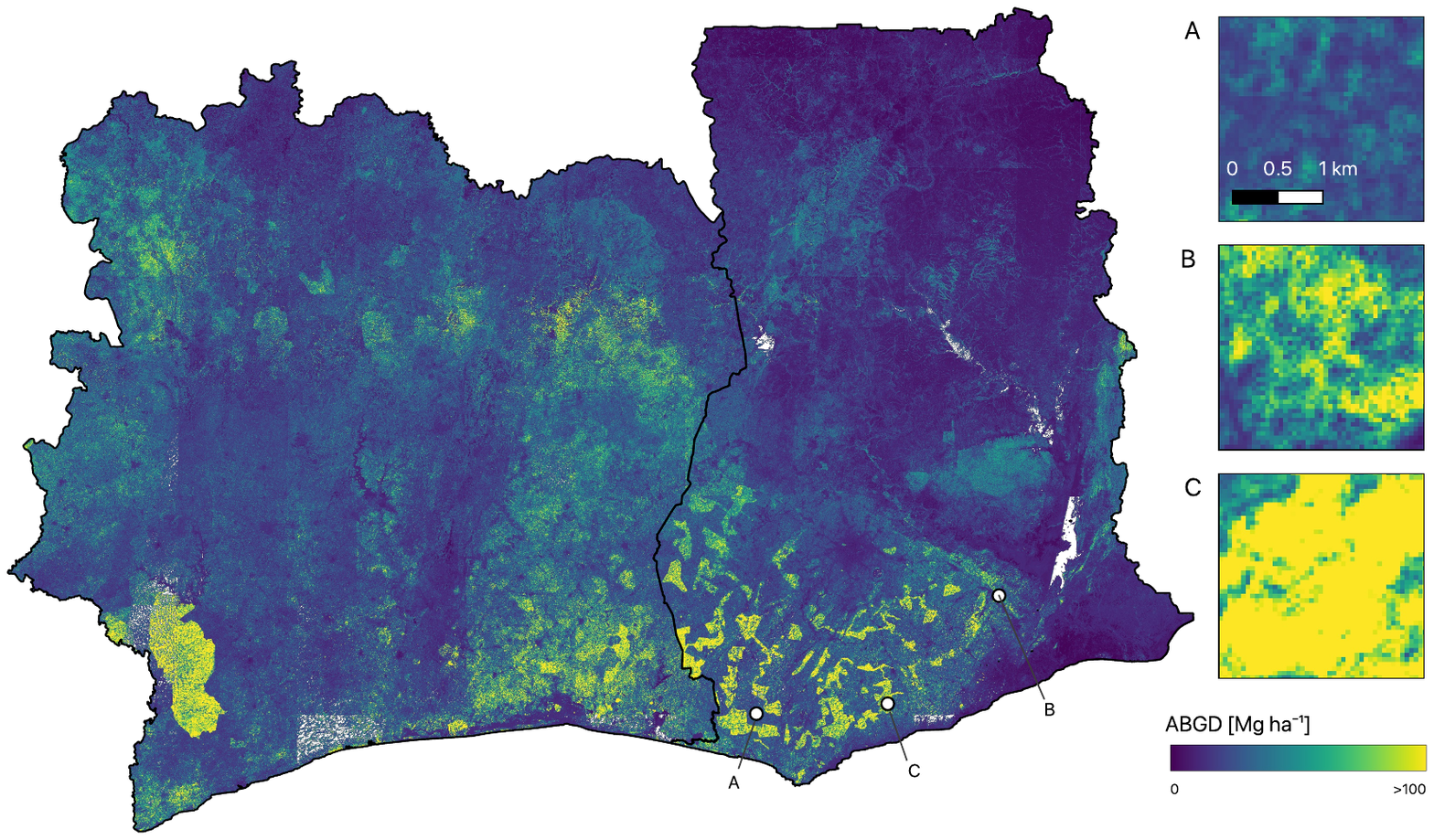}
    \caption{\textbf{Map of aboveground biomass density (AGBD) across Côte d'Ivoire and Ghana for the year 2022.} The AGBD estimates were derived from Sentinel-2 imagery at a ground sampling distance of $50 \times 50$\,m. Location A illustrates AGBD patterns in a cocoa monoculture in the Western Region of Ghana. Location B depicts an area of cocoa agroforestry in the Eastern Region, while Location C shows undisturbed forest in Kakum National Park in the Central Region of Ghana. White areas indicate pixels for which no input image was available, often due to persistent cloud cover. Base map boundaries were derived from the Global Administrative Areas database (GADM v4.1, \href{https://gadm.org/}{https://gadm.org}), used under academic license.}
    \label{fig:agbd_map}
\end{figure}

\begin{figure}
    \centering
    \includegraphics[width=0.5\textwidth,trim={5px 30px 5px 30px},clip]{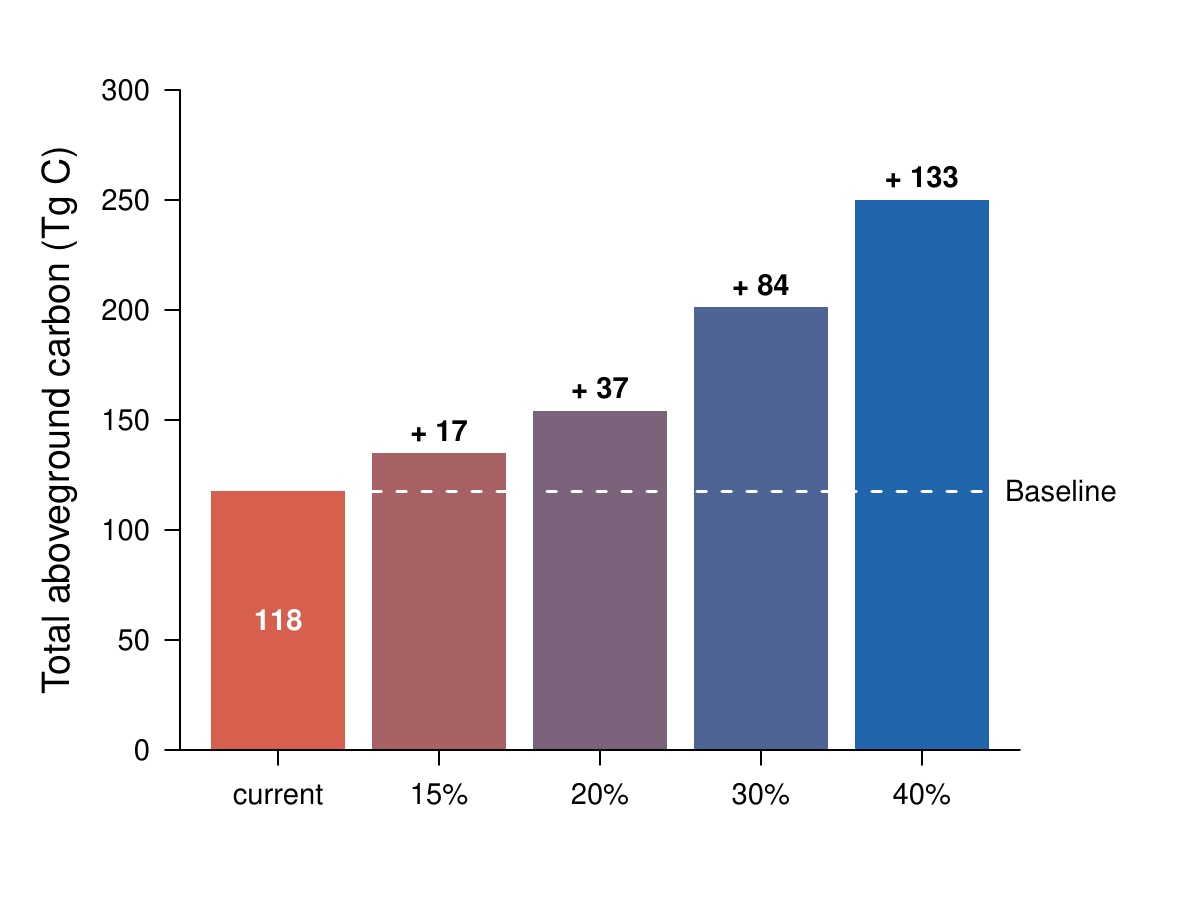}
    \vspace{-0.2em}
    \caption{\textbf{Current aboveground carbon stocks and climate mitigation potential of cocoa systems across Ghana and Côte d’Ivoire.} Bars represent aboveground carbon stocks under current conditions and projected scenarios with increased shade-tree cover (15\%, 20\%, 30\%, and 40\%). Numbers above the bars indicate the additional carbon stored compared to current levels. Due to the large number of estimates (over \(3 \times 10^{7}\) from our maps), the confidence intervals for current and projected values are too narrow to be visualized effectively. The 95\% Confidence Interval for current carbon stocks is $\pm 1.2 \times 10$\textsuperscript{-5}. For the future projections, the 95\% Highest Density Intervals (HDIs) are  $\pm 4.4 \times 10$\textsuperscript{-2}, $\pm 7.5 \times 10$\textsuperscript{-2}, $\pm 3.0 \times 10$\textsuperscript{-1}, and $\pm 7.0 \times 10$\textsuperscript{-1} million tonnes C for 15\%, 20\%, 30\%, and 40\% shade-tree cover adoption scenarios, respectively. The HDIs represent the range within which the means are statistically likely to reside.}
    \label{fig:C_potential}
\end{figure}

Estimates of carbon storage across cocoa growing areas are high relative to unforested areas (Fig. \ref{fig:agbd_map}), partly because cocoa itself is a medium-sized tree that also contributes to carbon sequestration. But because average shade-tree cover in cocoa growing areas is low (Fig. \ref{fig:shade_map}), estimates of existing carbon densities and even future potential carbon sequestration capacity in these systems pale in comparison to those in the small remaining patches of intact forests in the region (Fig.~\ref{fig:agbd_map}, Extended Data Fig.~\ref{fig:landuse_carbon_stocks}). Indeed, although the area under cocoa production is more than three and a half times larger than the area of remaining intact forests, it stores only  $\approx$40\% more carbon (Fig.~\ref{fig:agbd_map}, Extended Data Fig.~\ref{fig:landuse_carbon_stocks}). This is because of large differences in carbon density between these two systems (mean $\pm$ standard deviation; intact forest:  $36.6\pm 19.7$ tonnes C $ha^{-1}$; cocoa: $13.7\pm 6.9$ tonnes C $ha^{-1}$, Extended Data Fig.~\ref{fig:landuse_carbon_stocks}). Even if shade-tree cover levels were increased to 30\% across the entire cocoa growing region, the additional 84 million tonnes carbon (Fig.~\ref{fig:C_potential}) would represent only a little more than what is currently stored in the much smaller area of remaining intact forests (82.7 million tonnes carbon, Extended Data Fig.~\ref{fig:landuse_carbon_stocks}c). These numbers emphasize the critical importance of intact forests (or "avoided forest conversion"~\cite{griscom2017}) for climate-change mitigation, even before accounting for their contribution to biodiversity and other ecosystem services \cite{watson2018}. In this context, findings about the continuing contribution of cocoa production to deforestation, including in protected areas, are all the more concerning \cite{kalischek23natfood,Renier_2023,barima2016,ruf2015}.

\subsection*{Implications for implementation and monitoring}
Our findings highlight the enormous potential for climate-change mitigation offered by the widespread implementation of cocoa agroforestry across Ghana and Côte d’Ivoire. While not measured directly in our study, known positive relationships between shade-tree cover and climate adaptation \cite{blaser18,lin2007}, and shade-tree cover and biodiversity \cite{blaser18,bennett2022,bisseleua2009}, suggest significant additional sustainability benefits would also emerge, with limited or no loss of cocoa yield \cite{blaser18,andres2018,ramirez-argueta2022,bisseleua2009,asare2018,schroth2016c}. However, the effort required to achieve these positive outcomes should not be underestimated. The Cocoa and Forest Initiative (CFI), which is one of the most successful industry initiatives for combating deforestation and promoting sustainable cocoa production, resulted in 31.1 million shade trees planted across approximately 570,000 hectares (assuming planting density 54.6 trees ha\(^{-1}\)) between 2018 and 2022 (\href{https://worldcocoafoundation.org/news-and-resources/press-release/cocoa-forests-initiative-cfi-reports-progress-towards-ending-deforestation-through-cooperation-between-cocoa-sector-and-governments-of-cote-divoire-and-ghana}{CFI progress report 2023}, \href{https://worldcocoafoundation.org/storage/files/cfi-aggregate-action-plan-gh-022819.pdf}{CFI initial action plans for Ghana} and \href{https://worldcocoafoundation.org/storage/files/cfi-aggregate-action-plan-cdi-022819.pdf}{Côte d’Ivoire}). Despite this effort, this amounts to just 8.1\% of the cocoa growing area in this region \cite{kalischek23natfood}, and (assuming 30-40\% cover at maturity) would result in an increase average shade-tree cover across the region from 13.2\% to approximately 14.6-15.4\%. While not insignificant (Fig.~\ref{fig:C_potential}), this suggests that, in the short-term, such initiatives should be targeted to areas where maximum benefits can be achieved, such as regions particularly vulnerable to climate threats, or where other sustainability benefits such as supporting biodiversity can be most effectively achieved. More generally, the adoption of agroforestry by small-holder farmers faces several barriers, including limited short-term incentives, insecure land tenure, and the availability of forested fertile land that can be rapidly converted into new production areas \cite{kalischek23natfood,ruf2011}. Moreover, the climate benefits of tree planting hinge on permanence \cite{griscom2017,anderegg2020}, which can be undermined by economic pressures, governance failures, or climate extremes  \cite{ruf2011,unruh2008,reichstein2013,anderegg2020}. Clearly, realizing the large estimated benefits we describe will require an equally substantial and sustained effort in implementation. This effort must prioritize maximizing production within existing cultivated areas, in order to avoid further loss of intact forests that offer disproportionately high climate and biodiversity value. However, regardless of the nature and success of this effort, the maps and methods developed here provide a baseline and practical tools for monitoring agroforestry adoption and carbon sequestration over time, supporting sustainability reporting and guiding targeted interventions.

\subsection*{Advantages over previous work}
Our maps and methods offer several advantages compared to previous work. First, we use a state-of-the-art cocoa map\cite{kalischek23natfood} to mask the shade-cover map, ensuring that misidentification of land use does not cause areas of shade-tree cover in these systems to be over- or under-estimated. Second, our shade map predicts shade levels with low error ($\approx$5\%, Supplementary Fig.~\ref{fig:agbd_comp}a), and provides continuous shade levels, allowing users to employ and compare different thresholds between 'shaded' and 'unshaded' cocoa. Beyond compatibility with different specifications, the continuous scale increases reporting transparency, which is particularly important in the light of the ongoing debate about the definition of cocoa agroforestry, and varying targets set by individual companies and certification bodies. Third, our biomass map has higher resolution ($50\times50$ meters) than the widely used European Space Agency's (ESA) \href{https://climate.esa.int/en/projects/biomass/}{Climate Change Initiative (CCI) map} ($100\times100$ meters; Supplementary Methods). Moreover, we take particular care to avoid overestimating low above-ground biomass density (AGBD), which can often occur as a consequence of efforts to mitigate the underestimation of high biomass. By contrast, our approach minimizes these errors in cocoa-growing areas (Supplementary Fig. \ref{fig:agbd_comp}b), which is particularly important given that cocoa production systems are predominantly characterized by low biomass (Fig. \ref{fig:agbd_map}, Extended Data Fig.~\ref{fig:landuse_carbon_stocks}b). Finally, we aim for our project to support the implementation of zero-deforestation and carbon sequestration initiatives, particularly those relying on remote monitoring. As the demand grows for benchmarking commercial solutions, especially in relation to smallholder inclusion and carbon accounting, we provide our data, code, and maps openly. This transparency allows stakeholders to use current maps, but also to generate updated maps in the future, offering a cost-effective way for monitoring the long-term success of agroforestry interventions, and for meeting the demands of carbon accounting and sustainability reporting.

\subsection*{Limitations and future research}
We note three limitations to our work, with implications for future research. First, our methods required us to distinguish between shade trees and cocoa trees based on differences in their height (Methods). While our eight-meter cutoff minimizes the risk of misclassifying cocoa as shade trees, it could conceivably lead to the underestimation of shade-tree cover by excluding shade trees with canopies below eight meters in height. We think this risk is low given available estimates of shade-tree height in the study region \cite{blaser_2021}. Moreover, counter to this concern, our estimates for the total area of cocoa grown under 15\% shade cover is substantially larger than previous estimates (cf. land use map for Ghana, described above). It is important to note, however, that newly-planted shade trees typically require 5-10 years to reach a height of eight meters, and so our findings do not account for the potential influence of recent plantings on future shade-tree cover. Second, because our biomass estimates (AGBD) are based on the canopy height estimates of Lang et al.~\cite{lang2023high}, we inherit the overly smooth appearance of that map. As a result, the effective resolution of our AGBD map may be lower than its $50\times50$ meter ground sampling distance. We do not expect these effects to adversely affect our conclusions, but caution against quantitative analyses and interpretations at small scales. Finally, while we focus on shade-tree cover, our approach does not consider the full complexity of agroforestry systems, including species composition, structural diversity, or information on farm management, for example \cite{somarriba2024,blaser_2021}. We believe a focus on cover is justified because it is consistent with previous work and agroforestry recommendations, directly influences solar insolation to the crop, can be related to biomass, and can be estimated from satellite images as we have shown \cite{blaser18,zuidema2005}. However, using remote sensing to facilitate large-scale assessments of other important features of agroforests is an important area for future research. More generally, extending our methods, and applying them to other globally significant commodities that can be grown in agroforests may be highly beneficial.

\section*{Methods}
\label{Sec:methods}

\subsection*{Study region}
Our study focused on the cocoa-producing landscapes of Ghana and Côte d'Ivoire, which span 7.04 million hectares \cite{kalischek23natfood} and contribute roughly 60\% of global cocoa production \cite{FAO22}. Much of the region’s original closed tropical forest has been converted into a patchwork of land uses, including shifting cultivation, fallow systems, and perennial cropping systems—most notably cocoa. Cocoa farming in this region supports around two million farmers, most of whom manage small farms of three to five hectares \cite{hainmueller2011sustainable, bymolt2018demystifying} and earn less than one dollar per day \href{https://www.mightyearth.org/wp-content/uploads/2017/09/chocolates_dark_secret_english_web.pdf}{(Mighty Earth)}, even as cocoa contributes to between 10\% and 20\% of the countries' GDP.  Despite its economic importance, the region faces declining productivity due to soil degradation, disease pressure, and threats from climate change, which are predicted to limit production in the future \cite{schroth2016b,blaser18,franzen2007}. Furthermore, extensive deforestation has left little remaining primary forest \cite{kalischek23natfood,Renier_2023}, underscoring the growing importance of cocoa agroforests for both biodiversity conservation and carbon sequestration.

\subsection*{Estimating and mapping shade-tree cover}
We used supervised machine learning to estimate shade-tree cover in cocoa production systems across Ghana and Côte d’Ivoire from optical satellite images, calibrated with ground-truth data. Below we describe the satellite data, ground-truth data, machine learning algorithms, and methods used to generate the final maps of shade-tree cover. 

\textit{Satellite data.} We used satellite images from the Sentinel-2 constellation operated by the European Space Agency (https://sentinels.copernicus.eu/). Sentinel-2 satellites capture images of the earth across 13 spectral bands covering a wide range of the visible and near-infrared spectrum, at a spatial resolution between 10 and 60 meters per band, and at a revisit frequency of approximately 5 days for our study area. These images are ideal for our purpose as they are specifically collected to enable agro-environmental, forest, and vegetation mapping and monitoring. We collected atmospherically corrected, orthorectified surface reflectance images in 110$\times$110 km\textsuperscript{2} tiles across the study area. 

For model building and evaluation, we aimed to collect a set of training images that were, (i) cloud-free, (ii) taken close to the time of ground-truth data collection, and (iii) representative of different imaging conditions. To do this we first downloaded Sentinel-2 images in four time intervals (March 2020–May 2021; August 2021-December 2021; April 2022-July 2022; August 2022-December 2022). We downloaded all Sentinel-2 revisits of a given tile that fell within these intervals. Within each interval we sorted the tiles by the image-level cloud-cover attribute and retained up to $N=20$ of the least cloudy acquisitions. Some farms lie within multiple tiles, thus increasing the total number of available images for these farms. The set of images was then passed on to the training phase, during which we discarded images with clouds over the ground-truth farms (based on the per-pixel cloud cover map) and images taken more than 120 days from the date of the ground-truth data collection. From the remaining images we retained the ten images taken temporally-closest to the date of the ground-truth data collection. 

Our  approach minimizes the impact of temporal decorrelation caused by different acquisition dates of ground-truth and satellite imagery, and also minimizes the effect of phenology on shade-cover estimates by using data (satellite and ground-truth) collected predominantly during the growing season when shade-tree cover is expected to be highest. Our learning approach should account for any remaining variation in shade-cover during this time, after which cover estimates (i.e., maps of shade-tree cover) are generated based only on satellite images taken during the 2022 growing season (see below). We excluded remaining cloud-affected pixels from the final selection of ten images using the Sentinel-2 pixel-level cloud mask. Because our learning approach should accommodate variation caused by cloud shadows, we did not mask for this attribute.

\textit{Sampling design and methods for ground-truth data.} To calibrate the estimation from satellite images we collected direct field observations of shade-tree cover. We achieved this by capturing imagery with drones that we flew directly above 827 individual cocoa farms, distributed across four agro-climatic zones covering Ghana and Côte d’Ivoire. We focused our ground-truth sampling on farms that had been mapped in 2021 by two large cocoa trading companies. We developed a new agro-climatic zoning scheme to stratify our ground-truth sampling across cocoa-growing regions in Ghana and Côte d’Ivoire (see Supplementary Methods). In Ghana, we sampled 698 farms distributed according to a four-level stratified sampling design: cocoa farms within communities within administrative districts within four agro-climatic zones (Extended Data Fig.~\ref{fig:aez}, Supplementary Methods). In Côte d’Ivoire we collected ground-truth data at 60 additional farms distributed across three agro-climatic zones (Extended Data Fig.~\ref{fig:aez}). For effective ground-truthing of the satellite imagery, it was important that ground-truth data be collected on cocoa farms across the widest possible range of shade-tree cover levels that occur within the region. Because shade levels tended to be low in general, we sampled an additional 69 farms that we specifically targeted for their higher shade levels (Supplementary Methods).

For each farm, we flew a DJI Mavic 2 drone at a height of 80m to capture high-resolution images above the shade-tree canopy. All flights were planned, flown and processed using the cloud-based drone mapping software, DroneDeploy (Version 2.236.0). We used farm boundary polygons provided by cocoa trading companies to guide flight planning and ensure that each flight covered the area within or immediately surrounding the intended farm.

For our final estimates of shade-tree cover it was important to separate cover from shade trees from cover from cocoa trees. We did this using tree height to distinguishing between shade trees and cocoa trees. To do this we first photogrammetrically reconstructed a digital surface model (DSM) of the canopy, and a digital terrain model (DTM), and subtracted the two to obtain a canopy height map for each farm (Supplementary Methods). We then used an empirically-validated threshold height of eight meters to distinguish between shade trees and cocoa trees (Supplementary Methods). This threshold height was used to isolate (‘mask’) cocoa canopies from shade-tree canopies in our canopy height maps computed from the drone data. The resulting shade-cover maps served as local ground-truth reference data to train, validate, and test the machine learning model for satellite data, which was then deployed at country scale. All quality assessments of the resulting large-scale maps are also based on drone-derived reference data.

\textit{Machine-learning methods and shade-cover model.} We applied gradient boosting regression (GBR)~\cite{friedman2001greedy} to estimate shade-tree cover from satellite images across Ghana and Côte d’Ivoire. GBR is an ensemble machine learning algorithm, renowned for its efficacy in predictive modeling tasks. It operates in a fully supervised setting, using the per-farm shade cover maps derived from drone images as target values and Sentinel-2 L2A data as input. GBR learns an ensemble model in sequential fashion, such that each ensemble member further decreases the overall prediction error via an additive correction. We employed the Python implementation of GBR in scikit-learn (version 1.2.1, \url{https://scikit-learn.org/stable/}), with the number of decision trees set to 2000 and the robust Huber loss~\cite{huber64} as objective function. For training, we enumerated all farm polygons that form the training portion of the dataset and computed the ground truth shade cover per $10 \times 10$ meter Sentinel grid cell as the fraction of shade-tree pixels in the drone imagery that fall within that cell.

We used all twelve channels of the Sentinel-2 L2A product, at the 10 meter grid spacing defined by the blue, green, red and near-infra red channels. Lower resolution channels were upsampled to the same 10 meter grid with nearest neighbor interpolation. Besides raw spectral intensities we also computed a set of vegetation-related indices from literature (NDVI, GRVI, RVI, GNDVI and NDMI)~\cite{abderazak96} and appended them as additional input channels, for a total of 17 input values per pixel. For each pixel we then concatenated the values of all neighbours within a centred 5$\times$5 neighbourhood (i.e., 50$\times$50 meters on the ground). This approach avoids overly grainy results, and, assuming that differences between a centre pixel and its neighborhood are not random, allows us to use features from surrounding pixels to better predict cover for centre pixels. We also note that this neighbourhood size was empirically validated using independent validation sets in order to identify the neighbourhood that provided the best predictions. Ultimately there were 425 input features per pixel. These feature vectors were paired with the corresponding reference shade cover values, computed from the drone-based high-resolution maps.

Feature vectors and associated reference values were extracted for every pixel (i.e. densely) wherever ground-truth values were available. One third of those values was set aside as independent test set, on the remaining two thirds we trained GBR regressors with five-fold cross validation to identify optimal hyperparameters. A final model was trained on the entire dataset (excluding test set) and evaluated on the test set. All accuracy metrics in this manuscript were computed with that setting. 

\textit{Map of shade-tree cover} Using the trained GBR model, we estimated shade cover densely (i.e. per pixel) across the cocoa planted area of Ghana and Côte d’Ivoire, restricting predictions to areas mapped as cocoa-growing regions using the cocoa mask from Kalischek et al. (2023)\cite{kalischek23natfood}. This map distinguishes shaded and unshaded cocoa farms, including young, mature, and old cocoa farms, from surrounding landscape features. We note that, despite some misclassification of minor (in terms of area) landscape elements such as roads and scattered farmhouses, the cocoa classification method of Kalischek et al.\ (2023)\cite{kalischek23natfood} reliably identified cocoa farms regardless of their shade level, ensuring that both shaded and unshaded cocoa systems were included \cite{Moraiti2024}, which is a key requirement for meeting the goals of our study. To generate the shade-cover map, we used Sentinel-2 L2A images taken between April 2022 and November 2022, aligning with the growing season in this region. For each tile, images taken during this period were sorted quantitatively according to their cloud cover using the Sentinel-2 image-level cloud cover percentage attribute, after which the ten least cloud-affected images were selected. We excluded remaining cloud-affected pixels in these images from our analyses based on the Sentinel-2 pixel-level cloud mask.

All ten satellite images were processed into per-pixel feature vectors as derived above and mapped to per-pixel shade cover values with the GBR model, yielding ten shade cover maps. The ten estimates were averaged into a final shade cover map. Averaging across multiple acquisitions reduces the influence of remnant atmospheric effects on the final prediction. The final shade cover map of Ghana and Côte d’Ivoire (Fig.~\ref{fig:shade_map}), at 10 m ground sampling distance (GSD), was visualized in QGIS (v3.34.12\cite{QGIS}).

\subsection*{Estimating and mapping above-ground biomass density}
In a separate step we utilized a deep neural network to generate high-resolution estimates of above ground biomass density (AGBD) in the cocoa-producing regions of Ghana and Côte d’Ivoire. Below we describe the key datasets on which our estimates are based, and our deep learning model to generate the AGBD maps.

\textit{Data.} AGBD estimates were derived from two key datasets. The first one were sparse, L4A footprint-level AGBD estimates (25 m diameter) retrieved from spaceborne LiDAR (Light Detection and Ranging) with statistical models, as part of NASA’s Global Ecosystem Dynamics Investigation (GEDI)~\cite{duncanson2022aboveground}. GEDI offers AGBD estimates with good geographic coverage, but only at sparse footprint locations (the dense L4B product, accumulated from L4A, has 1 km GSD, too coarse for our purposes~\cite{saarela2018generalized}). By itself, the L4A product can therefore hardly be used to infer AGBD for the highly variable, small-scale cocoa farms distributed across our study area. Hence, we combined them with satellite-derived canopy height data (described further below). For our study, we acquired GEDI L4A AGBD measurements at individual 25 m diameter footprints distributed across the study area. Consistent with the timing of other data sources we used footprints acquired between April and November, 2022. We further filtered these to: 1) discard footprints that are deemed unreliable~\cite{lang2023high,duncanson2022aboveground,dubayah2022gedi,lanfranchi22}, and 2) only use data from GEDI's "power" beams, which are considered more reliable than the "coverage" beams for low biomass. After filtering, our AGBD dataset consisted of $\approx$2.1 million footprints with AGBD estimates, spread roughly evenly across the study area.

The second key dataset we used are raster maps of vegetation height, with associated per-pixel uncertainty values, covering the entire study area. To obtain these maps we applied the global canopy height retrieval model of Lang et al. (2023)~\cite{lang2023high} to the ten least cloudy Sentinel-2 images per tile over the year 2022 and averaged the 10 retrieval results. Two properties make this dataset useful for our task of dense AGBD retrieval. First, canopy height is known to strongly correlate with AGBD~ \cite{lefsky05,asner12}, meaning that it is a suitable predictor where direct AGBD measurements are absent. Previous work has shown that predicting AGBD directly from a dense canopy height map gives satisfactory results\cite{lanfranchi22}. Second, in contrast to the L4A product, the canopy height maps offer dense, wall-to-wall coverage, with only a few pixels rendered invalid due to persistent cloud cover. This extensive coverage can help interpolate between GEDI footprints, effectively 'filling in the gaps' in that data. More generally, it is computationally more lightweight to directly lift the existing canopy height map to an AGBD map instead of estimating AGBD values from Sentinel-2 images, especially if local or regional maps are needed as in our case.

\textit{Deep-learning methods and biomass model.} We used a convolutional neural network (CNN) to estimate AGBD across Ghana and Côte d’Ivoire. CNNs are a type of deep learning model designed to interpret spatially organized data, such as image pixels arranged in a grid They apply a series of learnable filters to the input image, which are adept at capturing spatial patterns in a hierarchical fashion. For our work, we trained a CNN to densely predict AGBD at 50 m grid spacing, using as input 150$\times$150 m\textsuperscript{2} patches of canopy height (sampled at 10 m grid spacing) and a periodic encoding of the path centre latitude. The latter makes it possible to also learn biomass patterns governed by the latitudinal location of the patch. We note that we use larger grid spacing (50$\times$50 m\textsuperscript{2}) and context window (150$\times$150 m\textsuperscript{2}) for the AGBD model than our cover model because we use different analytical approaches (CNN vs. GBR) and different types of input data to model these different variables, and because empirical cross validation identified the context windows with highest predictive power in each case. 

As reference data to train the CNN, we use the GEDI L4A AGBD data, described above. Note that we rely on the GEDI L4A AGBD product for model training, validation, and testing. As output, we estimate an AGBD value and its standard deviation per grid cell, assuming an approximately Gaussian error distribution. The predicted, calibrated standard deviation is of particular value in our setting as a measure of trustworthiness, where AGBD estimates with higher predictive standard deviation are less reliable.

Following standard practice we train our model in an iterative manner using stochastic gradient descent and back-propagation, where model parameters are updated at every step to increase the likelihood of the ground truth targets under the model’s output distribution. Each sample's contribution to the overall error is reweighted according to the inverse square root of the frequency of the corresponding biomass value in the overall dataset, to reduce the bias towards the predominant AGBD values. We divide the dataset into 20 equidistant bins from 0 to 200 tonnes ha\(^{-1}\) and assign the same frequency to all samples in each bin. Our CNN has six layers, with kernel size 3$\times$3 and channel depths \{16, 32, 64, 128, 128, 128\}. We use the Adam optimizer~\cite{KingmaB14} with a base learning rate (step size) of $10^{-5}$. To estimate the predictive standard deviation, and to boost performance, we train an ensemble of five models with different random initialisations. Before training, the data is split into training, validation and testing portions, by dividing each Sentinel-2 tile into five longitudinal stripes of width 20 km and assigning three stripes to training, one to validation and one to testing. Optimisation was run until the validation performance no longer improved, and model performance was evaluated on the independent test set.

\textit{Map of aboveground biomass} We generated a high-resolution (50$\times$50 m\textsuperscript{2}) map of AGBD across Ghana and Côte d’Ivoire. This map includes all cocoa growing areas and non-cocoa growing areas. To generate this map, we fed all 110 km × 110 km canopy height (and uncertainty) tiles of our study area through the trained CNN model. The resulting AGBD map Fig.~\ref{fig:agbd_map}, at 50 m ground GSD, was visualized in QGIS (v3.34.12\cite{QGIS}).

\textit{Comparison of cocoa and forest areas.} To compare area, carbon density, and total carbon storage across the three land-use classes — \emph{cocoa}, \emph{disturbed forest}, and \emph{undisturbed forest} — we used our 2022 shade-cover map to delineate cocoa-growing areas. Forest areas were derived from the Tropical Moist Forest (TMF) map produced by the European Commission’s Joint Research Centre (JRC), which depicts disturbed and undisturbed tropical moist forests for the year 2022 \cite{vancutsem2021}. Both maps were bilinearly resampled to match the resolution of our AGBD map (50$\times$50 m\textsuperscript{2}), enabling us to filter it at its native resolution and calculate the area, mean and standard deviation of AGBD, as well as total aboveground biomass and 95\% confidence intervals (CIs) for each land-use class. Note that areas mapped as cocoa in our shade-cover map were excluded from the forest map to ensure the land-use classes remained distinct. Biomass was converted to carbon assuming 47\% of above-ground biomass is carbon, following Martin et al.~(2011) \cite{martin2011}.

\subsection*{Climate-change mitigation scenarios}
To quantify how different levels of future potential shade-tree cover translates to carbon sequestration, one must first quantify the relationship between shade-tree cover and biomass. This was done by converting the AGBD estimates at each pixel into estimates of total aboveground biomass (AGB) per pixel. To ensure consistent spatial alignment and enable accurate pixel-to-pixel comparisons, we bilinearly resampled the shade map (10$\times$10 m\textsuperscript{2}) to match the resolution of the biomass map (50$\times$50 m\textsuperscript{2}). We then quantified the relationship between shade-tree cover and total biomass with Bayesian linear regression, using our pixel-level estimates of shade-tree cover as the predictor variable, and our pixel-level estimates of total aboveground biomass as the response variable (Supplementary Methods). 

We used our empirical relationship between shade-tree cover and above-ground biomass stocks (Fig.~\ref{fig:hexbin_cover_biomass}) to estimate the amount of sequestered carbon under current conditions, and under four potential future levels of shade-tree cover (15\%, 20\%, 30\% and 40\%). These four threshold levels are indicative and were chosen because they correspond to \href{https://cdm.unfccc.int/DNA/cdf/files/2008/1706_ghana.pdf}{Ghana’s national definition of forest}, and to various industry recommendations for shade-tree cover in cocoa agroforests (\href{https://voicenetwork.cc/wp-content/uploads/2020/07/200706-Cocoa-Barometer-Agroforestry-Consultation-Paper.pdf}{Sanial et al. (2020)}).

We will describe our methods for a threshold level of cover of 30\%, noting that we used the same procedure for all threshold levels. For any one hypothetical pixel, we first use our fitted regression model to estimate the biomass associated with a cover level of 30\%. This was done by randomly drawing 100 samples of paired intercept and slope estimates from the posterior distributions of our fitted regression model, and then using each of these unique parameter combinations to generate a posterior distribution of biomass for a hypothetical pixel with 30\% cover. We then multiplied this posterior distribution by the total number of pixels in our cover map that have less than 30\% shade-tree cover, which gives a posterior distribution of the biomass that would be added by increasing the shade cover of those pixels to 30\%. We then added the total biomass of all pixels already at or above the threshold level by extracting these values directly from our biomass map. Our final biomass estimate is a posterior distribution of total biomass across all locations where shade tree cover is increased to 30\% wherever it is below that threshold, and remains unchanged for all locations where the threshold is currently satisfied. From this distribution, we extracted both the mean total carbon and a 95\% Highest Density Interval (HDI), representing the range within which the mean biomass is statistically most likely to reside. Biomass was converted to carbon assuming that carbon represents 47\% of the above-ground biomass, following Martin et al.~(2011) \cite{martin2011}. The same procedure was repeated for all threshold levels. Additional details of our carbon sequestration calculations are described in the Supplementary Methods. 

\section*{Data availability}
All key datasets used in this study are publicly available. The final maps of shade-tree cover, canopy height, and aboveground biomass density can be explored interactively and downloaded via Google Earth Engine: \url{https://albecker.users.earthengine.app/view/agroforestry}. The drone-derived ground-truth data for shade-tree cover are available via UQ eSpace  \url{https://doi.org/10.48610/dda018c}\cite{blaser-hart2025cocoa}. Sentinel-2 satellite imagery is available from the Copernicus Open Access Hub: \url{https://sentinels.copernicus.eu}. Aboveground biomass reference data were obtained from NASA GEDI L4A (2022), available at \url{https://doi.org/10.3334/ORNLDAAC/2056}. Canopy height data were generated using the model of Lang et al. (2023)\cite{lang2023high}. The cocoa-growing area mask used in this study was obtained from Kalischek et al. (2023)\cite{kalischek23natfood} and used as provided. Forest classifications were based on the JRC Tropical Moist Forest (TMF) map: \url{https://forobs.jrc.ec.europa.eu/TMF/}. Climatic variables were derived from the CHIRTS-daily temperature dataset and the CHIRPS v2.0 precipitation dataset: \url{https://www.chc.ucsb.edu/data}. Potential evapotranspiration (PET) data were obtained from the Global Aridity and PET Database: \url{https://cgiarcsi.community/data/global-aridity-and-pet-database/}. Administrative boundaries were sourced from GADM v4.1: \url{https://gadm.org}. The greenhouse gas emission estimates for cocoa production were provided by Quantis using the World Food LCA Database, which is not publicly available; however, the exact values used in this study are reported in the manuscript. Farm boundary polygons used to guide drone flight planning were provided by cocoa trading companies and are not publicly available due to confidentiality agreements.

\section*{Code availability}
The code used to estimate shade-tree cover and above-ground biomass density is available on GitHub: \url{https://github.com/prs-eth/agroforestry}. This repository includes code for pre-processing satellite imagery, training the gradient boosting and deep learning models (implemented in scikit-learn and PyTorch), and generating the final maps. The code used to calculate the climate-change mitigation scenarios is available via UQ eSpace \url{https://doi.org/10.48610/dda018c}\cite{blaser-hart2025cocoa}.

\section*{Corresponding author}
Correspondence to \href{mailto:w.hart@uq.edu.au}{Wilma J. Blaser-Hart}.

\section*{Acknowledgements}
We thank R. Tetteh, J. Afele, E. Nimo, M. Yombu, V. Agumenu, and Hammond for their exceptional efforts in visiting hundreds of cocoa farms to collect the ground truth data in Ghana. Our thanks also extend to Barry Callebaut and Olam Food Ingredients (ofi) for their invaluable support in making the ground campaign possible. Special thanks to M. Gilmour, S. Ankamah, C. Parra Paitan, O. Nkuah, E. Prempeh, R. Seidu, B. Karibu, S. Adusei, S. Dodzie, E. Obiri Yeboah, S. Larbie, and D. Forson and their ground-level field teams for their support in helping us gain access to recently mapped cocoa farms . We are grateful to the German Development Agency (GIZ) for their collaboration, which enabled us to extend our ground truth sampling to Côte D'Ivoire. Special thanks to H. Walz, A. Bio, P. Ripplinger, and M. Pallauf. We also appreciate the support of P. Kouakou in processing high-resolution drone images for Côte d'Ivoire. Thanks to Dr. Alexi Ernstoff, Vincent Rossi, Cecile Guignard, and Tereza Levova from Quantis for their assistance with analysis of annual carbon emissions from cocoa production. This project received funding from the Lindt Cocoa Foundation (W.J.B., S.P.H, J.D.W), the 2019–2020 BiodivERsA joint call for research proposals under the BiodivClim ERA-Net COFUND program, and with the funding organization of the Swiss National Science Foundation (FNRS under Grant n°PINT MULTI/BEJ—R.8002.20; R.D.G., C.B., J.D.W. and W.J.B.), the Joint Cocoa Research Fund of CAOBISCO and ECA (W.J.B., S.P.H, J.D.W), and the Queensland Government under the Women's Research Assistance Program (WRAP194-2019RD1; W.J.B.). We thank the ESA Copernicus programme for its commitment to open data access, which made this research possible.

\section*{Author contributions}

W.J.B., S.P.H., and J.D.W.\ conceived the research question and designed the study. W.J.B.\ planned the fieldwork, with W.J.B.\ and E.D.\ overseeing its execution. A.B.\ developed the code with guidance from J.D.W.\ and K.S. W.J.B., A.B., and S.P.H.\ analyzed the results. C.B.\ and F.C.\ created the maps of agro-climatic zones. W.J.B.\ and S.P.H.\ led the writing, with contributions from all authors.

\section*{Competing interests}
The authors declare no competing interests.



\bibliography{main}

\begin{thebibliography}{10}

\bibitem{breiman2001}
Leo Breiman.
\newblock Random forests.
\newblock {\em Machine Learning}, 45(1):5--32, 2001.

\bibitem{Kamath2024}
Vignesh Kamath, Marieke Sassen, Andy Arnell, Arnout {van Soesbergen}, and Christian Bunn.
\newblock Identifying areas where biodiversity is at risk from potential cocoa expansion in the congo basin.
\newblock {\em Agriculture, Ecosystems \& Environment}, 376:109216, 2024.

\bibitem{kalischek23natfood}
N.~Kalischek, N.~Lang, C.~Renier, R.~C. Daudt, T.~Addoah, W.~Thompson, W.~J. Blaser-Hart, R.~Garrett, K.~Schindler, and J.~D. Wegner.
\newblock Cocoa plantations are associated with deforestation in {C{\^o}te d'Ivoire} and {Ghana}.
\newblock {\em Nature Food}, 4:384--393, 2023.

\bibitem{funk2019}
Chris Funk, Pete Peterson, Seth Peterson, Shraddhanand Shukla, Frank Davenport, Joel Michaelsen, Kenneth~R. Knapp, Martin Landsfeld, Gregory Husak, Laura Harrison, James Rowland, Michael Budde, Alex Meiburg, Tufa Dinku, Diego Pedreros, and Nicholas Mata.
\newblock A high-resolution 1983–2016 {Tmax} climate data record based on infrared temperatures and stations by the {Climate} {Hazard} {Center}.
\newblock {\em Journal of Climate}, 32(17):5639--5658, 2019.

\bibitem{funk2014}
Chris~C. Funk, Pete~J. Peterson, Martin~F. Landsfeld, Diego~H. Pedreros, James~P. Verdin, James~D. Rowland, Bo~E. Romero, Gregory~J. Husak, Joel~C. Michaelsen, and Andrew~P. Verdin.
\newblock A quasi-global precipitation time series for drought monitoring.
\newblock Report 832, United States Geological Survey (USGS), 2014.

\bibitem{trabucco2018}
Antonio Trabucco and Robert~J. Zomer.
\newblock Global aridity index and potential evapotranspiration ({ET0}) climate database v2.
\newblock CGIAR Consort Spat Inf 10: m9, 2018.

\bibitem{Allan1998}
Richard Allan, L.~Pereira, and Martin Smith.
\newblock Crop evapotranspiration - {Guidelines} for computing crop water requirements-{FAO} irrigation and drainage paper 56.
\newblock FAO, Rome 300 (9): D05109, 1998.

\bibitem{liaw2002}
Andy Liaw and Matthew Wiener.
\newblock Classification and regression by {randomForest}.
\newblock R News 2 (3): 18–22, 2002.

\bibitem{R2021}
{R Core Team}.
\newblock {\em R: A Language and Environment for Statistical Computing}.
\newblock R Foundation for Statistical Computing, Vienna, Austria, 2021.

\bibitem{shi2006}
Tao Shi and Steve Horvath.
\newblock Unsupervised learning with random forest predictors.
\newblock {\em Journal of Computational and Graphical Statistics}, 15(1):118--138, 2006.

\bibitem{VanDerWal2009}
Jeremy VanDerWal, Luke~P. Shoo, Catherine Graham, and Stephen~E. Williams.
\newblock Selecting pseudo-absence data for presence-only distribution modeling: How far should you stray from what you know?
\newblock {\em Ecological Modelling}, 220(4):589--594, 2009.

\bibitem{bunn19}
Christian Bunn, Peter L\"aderach, Amos Quaye, Sander Muilerman, Martin~R.A. Noponen, and Mark Lundy.
\newblock Recommendation domains to scale out climate change adaptation in cocoa production in {Ghana}.
\newblock {\em Climate Services}, 16:100123, 2019.

\bibitem{blaser_2021}
W.~J. Blaser-Hart, S.~P. Hart, J.~Oppong, D.~Kyereh, E.~Yeboah, and J.~Six.
\newblock The effectiveness of cocoa agroforests depends on shade-tree canopy height.
\newblock {\em Agriculture, Ecosystems \& Environment}, 322:107676, 2021.

\bibitem{bürkner2017}
Paul-Christian Bürkner.
\newblock brms: An {R} package for bayesian multilevel models using stan.
\newblock {\em Journal of Statistical Software}, 80(1):1 -- 28, 2017.

\bibitem{nair2009}
P.~K.~Ramachandran Nair, B.~Mohan Kumar, and Vimala~D. Nair.
\newblock Agroforestry as a strategy for carbon sequestration.
\newblock {\em Journal of Plant Nutrition and Soil Science}, 172(1):10 – 23, 2009.
\newblock Cited by: 670.

\bibitem{somarriba2013}
Eduardo Somarriba, Rolando Cerda, Luis Orozco, Miguel Cifuentes, Hector Davila, Tania Espin, Henry Mavisoy, Guadalupe Avila, Estefany Alvarado, Veronica Poveda, Carlos Astorga, Eduardo Say, and Olivier Deheuvels.
\newblock Carbon stocks and cocoa yields in agroforestry systems of central america.
\newblock {\em AGRICULTURE ECOSYSTEMS \& ENVIRONMENT}, 173:46--57, JUL 1 2013.

\bibitem{Quantis2024}
{Quantis}.
\newblock {WFLDB – World Food LCA database, version 3.10, “Cocoa beans, sun-dried, agroforestry, at farm - without carbon capture in shade trees /GH U \& CI U” datasets}, 2024.

\bibitem{FAO22}
{Food and Agriculture Organization of the United Nations}.
\newblock Cocoa beans production. {FAOSTAT}.
\newblock Technical report, FAO, 2022.
\newblock Accessed: 2024-05-02.

\end{thebibliography}


\begin{thebibliography}{10}
\urlstyle{rm}
\expandafter\ifx\csname url\endcsname\relax
  \def\url#1{\texttt{#1}}\fi
\expandafter\ifx\csname urlprefix\endcsname\relax\def\urlprefix{URL }\fi
\expandafter\ifx\csname doiprefix\endcsname\relax\def\doiprefix{DOI: }\fi
\providecommand{\bibinfo}[2]{#2}
\providecommand{\eprint}[2][]{\url{#2}}

\bibitem{blaser18}
\bibinfo{author}{Blaser, W.~J.} \emph{et~al.}
\newblock \bibinfo{journal}{\bibinfo{title}{Climate-smart sustainable agriculture in low-to-intermediate shade agroforests}}.
\newblock {\emph{\JournalTitle{Nature Sustainability}}} \textbf{\bibinfo{volume}{1}}, \bibinfo{pages}{234--239}, \doiprefix\url{10.1038/s41893-018-0062-8} (\bibinfo{year}{2018}).

\bibitem{griscom2017}
\bibinfo{author}{Griscom, B.~W.} \emph{et~al.}
\newblock \bibinfo{journal}{\bibinfo{title}{Natural climate solutions}}.
\newblock {\emph{\JournalTitle{Proceedings of the National Academy of Sciences of the United States of America}}} \textbf{\bibinfo{volume}{114}}, \bibinfo{pages}{11645--11650}, \doiprefix\url{10.1073/pnas.1710465114} (\bibinfo{year}{2017}).

\bibitem{bennett2022}
\bibinfo{author}{Bennett, R.~E.}, \bibinfo{author}{Sillett, T.~S.}, \bibinfo{author}{Rice, R.~A.} \& \bibinfo{author}{Marra, P.~P.}
\newblock \bibinfo{journal}{\bibinfo{title}{Impact of cocoa agricultural intensification on bird diversity and community composition}}.
\newblock {\emph{\JournalTitle{Conservation Biology}}} \textbf{\bibinfo{volume}{36}}, \doiprefix\url{10.1111/cobi.13779} (\bibinfo{year}{2022}).

\bibitem{terasaki2023}
\bibinfo{author}{Terasaki~Hart, D.~E.} \emph{et~al.}
\newblock \bibinfo{journal}{\bibinfo{title}{Priority science can accelerate agroforestry as a natural climate solution}}.
\newblock {\emph{\JournalTitle{Nature Climate Change}}} \textbf{\bibinfo{volume}{13}}, \bibinfo{pages}{1179--1190}, \doiprefix\url{10.1038/s41558-023-01810-5} (\bibinfo{year}{2023}).

\bibitem{phalan2011}
\bibinfo{author}{Phalan, B.}, \bibinfo{author}{Onial, M.}, \bibinfo{author}{Balmford, A.} \& \bibinfo{author}{Green, R.~E.}
\newblock \bibinfo{journal}{\bibinfo{title}{Reconciling food production and biodiversity conservation: Land sharing and land sparing compared}}.
\newblock {\emph{\JournalTitle{Science}}} \textbf{\bibinfo{volume}{333}}, \bibinfo{pages}{1289--1291}, \doiprefix\url{10.1126/science.1208742} (\bibinfo{year}{2011}).

\bibitem{balmford2018}
\bibinfo{author}{Balmford, A.} \emph{et~al.}
\newblock \bibinfo{journal}{\bibinfo{title}{The environmental costs and benefits of high-yield farming}}.
\newblock {\emph{\JournalTitle{Nature Sustainability}}} \textbf{\bibinfo{volume}{1}}, \bibinfo{pages}{477--485}, \doiprefix\url{10.1038/s41893-018-0138-5} (\bibinfo{year}{2018}).

\bibitem{sills2019}
\bibinfo{author}{Sills, J.} \emph{et~al.}
\newblock \bibinfo{journal}{\bibinfo{title}{Forest restoration: Expanding agriculture}}.
\newblock {\emph{\JournalTitle{Science}}} \textbf{\bibinfo{volume}{366}}, \bibinfo{pages}{316--317}, \doiprefix\url{doi:10.1126/science.aaz0705} (\bibinfo{year}{2019}).

\bibitem{veldman2019}
\bibinfo{author}{Veldman, J.~W.} \emph{et~al.}
\newblock \bibinfo{journal}{\bibinfo{title}{Comment on “the global tree restoration potential”}}.
\newblock {\emph{\JournalTitle{Science}}} \textbf{\bibinfo{volume}{366}}, \bibinfo{pages}{eaay7976}, \doiprefix\url{doi:10.1126/science.aay7976} (\bibinfo{year}{2019}).

\bibitem{bastin2019}
\bibinfo{author}{Bastin, J.~F.} \emph{et~al.}
\newblock \bibinfo{journal}{\bibinfo{title}{Forest restoration: Transformative trees-response}}.
\newblock {\emph{\JournalTitle{Science}}} \textbf{\bibinfo{volume}{366}}, \bibinfo{pages}{317}, \doiprefix\url{10.1126/science.aaz2148} (\bibinfo{year}{2019}).

\bibitem{rosenstock2019}
\bibinfo{author}{Rosenstock, T.~S.} \emph{et~al.}
\newblock \bibinfo{journal}{\bibinfo{title}{Making trees count: Measurement and reporting of agroforestry in unfccc national communications of non-annex i countries}}.
\newblock {\emph{\JournalTitle{Agriculture, Ecosystems \& Environment}}} \textbf{\bibinfo{volume}{284}}, \bibinfo{pages}{106569}, \doiprefix\url{https://doi.org/10.1016/j.agee.2019.106569} (\bibinfo{year}{2019}).

\bibitem{brancalion2020}
\bibinfo{author}{Brancalion, P. H.~S.} \& \bibinfo{author}{Holl, K.~D.}
\newblock \bibinfo{journal}{\bibinfo{title}{Guidance for successful tree planting initiatives}}.
\newblock {\emph{\JournalTitle{Journal of Applied Ecology}}} \textbf{\bibinfo{volume}{57}}, \bibinfo{pages}{2349--2361}, \doiprefix\url{https://doi.org/10.1111/1365-2664.13725} (\bibinfo{year}{2020}).

\bibitem{holl2020}
\bibinfo{author}{Holl, K.~D.} \& \bibinfo{author}{Brancalion, P. H.~S.}
\newblock \bibinfo{journal}{\bibinfo{title}{Tree planting is not a simple solution}}.
\newblock {\emph{\JournalTitle{Science}}} \textbf{\bibinfo{volume}{368}}, \bibinfo{pages}{580--581}, \doiprefix\url{10.1126/science.aba8232} (\bibinfo{year}{2020}).

\bibitem{FAO22}
\bibinfo{author}{{Food and Agriculture Organization of the United Nations}}.
\newblock \bibinfo{title}{Cocoa beans production. {FAOSTAT}}.
\newblock \bibinfo{type}{Tech. Rep.}, \bibinfo{institution}{FAO} (\bibinfo{year}{2022}).
\newblock \bibinfo{note}{Accessed: 2024-05-02}.

\bibitem{kalischek23natfood}
\bibinfo{author}{Kalischek, N.} \emph{et~al.}
\newblock \bibinfo{journal}{\bibinfo{title}{Cocoa plantations are associated with deforestation in {C{\^o}te d'Ivoire} and {Ghana}}}.
\newblock {\emph{\JournalTitle{Nature Food}}} \textbf{\bibinfo{volume}{4}}, \bibinfo{pages}{384--393} (\bibinfo{year}{2023}).

\bibitem{Renier_2023}
\bibinfo{author}{Renier, C.} \emph{et~al.}
\newblock \bibinfo{journal}{\bibinfo{title}{Transparency, traceability and deforestation in the ivorian cocoa supply chain}}.
\newblock {\emph{\JournalTitle{Environmental Research Letters}}} \textbf{\bibinfo{volume}{18}}, \bibinfo{pages}{024030}, \doiprefix\url{10.1088/1748-9326/acad8e} (\bibinfo{year}{2023}).

\bibitem{barima2016}
\bibinfo{author}{Barima, Y. S.~S.} \emph{et~al.}
\newblock \bibinfo{journal}{\bibinfo{title}{Cocoa crops are destroying the forest reserves of the classified forest of {Haut-Sassandra} ({Ivory} {Coast})}}.
\newblock {\emph{\JournalTitle{Global Ecology and Conservation}}} \textbf{\bibinfo{volume}{8}}, \bibinfo{pages}{85--98}, \doiprefix\url{https://doi.org/10.1016/j.gecco.2016.08.009} (\bibinfo{year}{2016}).

\bibitem{poore2018}
\bibinfo{author}{Poore, J.} \& \bibinfo{author}{Nemecek, T.}
\newblock \bibinfo{journal}{\bibinfo{title}{Reducing food’s environmental impacts through producers and consumers}}.
\newblock {\emph{\JournalTitle{Science}}} \textbf{\bibinfo{volume}{360}}, \bibinfo{pages}{987--992}, \doiprefix\url{10.1126/science.aaq0216} (\bibinfo{year}{2018}).

\bibitem{andres2018}
\bibinfo{author}{Andres, C.} \emph{et~al.}
\newblock \bibinfo{journal}{\bibinfo{title}{Agroforestry systems can mitigate the severity of cocoa swollen shoot virus disease}}.
\newblock {\emph{\JournalTitle{Agriculture, Ecosystems \& Environment}}} \textbf{\bibinfo{volume}{252}}, \bibinfo{pages}{83--92}, \doiprefix\url{10.1016/j.agee.2017.09.031} (\bibinfo{year}{2018}).

\bibitem{clough2011}
\bibinfo{author}{Clough, Y.} \emph{et~al.}
\newblock \bibinfo{journal}{\bibinfo{title}{Combining high biodiversity with high yields in tropical agroforests}}.
\newblock {\emph{\JournalTitle{Proceedings of the National Academy of Sciences of the United States of America}}} \textbf{\bibinfo{volume}{108}}, \bibinfo{pages}{8311--8316}, \doiprefix\url{10.1073/pnas.1016799108} (\bibinfo{year}{2011}).

\bibitem{ramirez-argueta2022}
\bibinfo{author}{Ramírez-Argueta, O.} \emph{et~al.}
\newblock \bibinfo{journal}{\bibinfo{title}{Timber growth, cacao yields, and financial revenues in a long-term experiment of cacao agroforestry systems in northern honduras}}.
\newblock {\emph{\JournalTitle{Frontiers in Sustainable Food Systems}}} \textbf{\bibinfo{volume}{6}}, \doiprefix\url{10.3389/fsufs.2022.941743} (\bibinfo{year}{2022}).

\bibitem{bisseleua2009}
\bibinfo{author}{Bisseleua, D. H.~B.}, \bibinfo{author}{Missoup, A.~D.} \& \bibinfo{author}{Vidal, S.}
\newblock \bibinfo{journal}{\bibinfo{title}{Biodiversity conservation, ecosystem functioning, and economic incentives under cocoa agroforestry intensification}}.
\newblock {\emph{\JournalTitle{Conservation Biology}}} \textbf{\bibinfo{volume}{23}}, \bibinfo{pages}{1176--1184}, \doiprefix\url{10.1111/j.1523-1739.2009.01220.x} (\bibinfo{year}{2009}).

\bibitem{schroth2016c}
\bibinfo{author}{Schroth, G.} \emph{et~al.}
\newblock \bibinfo{journal}{\bibinfo{title}{{Climate friendliness of cocoa agroforests is compatible with productivity increase}}}.
\newblock {\emph{\JournalTitle{Mitigation and Adaptation Strategies for Global Change}}} \textbf{\bibinfo{volume}{21}}, \bibinfo{pages}{67--80}, \doiprefix\url{10.1007/s11027-014-9570-7} (\bibinfo{year}{2016}).

\bibitem{carodenuto2021}
\bibinfo{author}{Carodenuto, S.} \& \bibinfo{author}{Buluran, M.}
\newblock \bibinfo{journal}{\bibinfo{title}{The effect of supply chain position on zero-deforestation commitments: evidence from the cocoa industry}}.
\newblock {\emph{\JournalTitle{Journal of Environmental Policy \& Planning}}} \textbf{\bibinfo{volume}{23}}, \bibinfo{pages}{716--731}, \doiprefix\url{10.1080/1523908X.2021.1910020} (\bibinfo{year}{2021}).
\newblock \eprint{https://doi.org/10.1080/1523908X.2021.1910020}.

\bibitem{lang2023high}
\bibinfo{author}{Lang, N.}, \bibinfo{author}{Jetz, W.}, \bibinfo{author}{Schindler, K.} \& \bibinfo{author}{Wegner, J.~D.}
\newblock \bibinfo{journal}{\bibinfo{title}{A high-resolution canopy height model of the {Earth}}}.
\newblock {\emph{\JournalTitle{Nature Ecology \& Evolution}}} \bibinfo{pages}{1--12} (\bibinfo{year}{2023}).

\bibitem{knudsen2015}
\bibinfo{author}{Knudsen, M.~H.} \& \bibinfo{author}{Agergaard, J.}
\newblock \bibinfo{journal}{\bibinfo{title}{Ghana's cocoa frontier in transition: the role of migration and livelihood diversification}}.
\newblock {\emph{\JournalTitle{Geografiska Annaler: Series B, Human Geography}}} \textbf{\bibinfo{volume}{97}}, \bibinfo{pages}{325--342}, \doiprefix\url{10.1111/geob.12084} (\bibinfo{year}{2015}).
\newblock \eprint{https://doi.org/10.1111/geob.12084}.

\bibitem{thompson2022}
\bibinfo{author}{Thompson, W.} \emph{et~al.}
\newblock \bibinfo{journal}{\bibinfo{title}{Can sustainability certification enhance the climate resilience of smallholder farmers? the case of {Ghanaian} cocoa}}.
\newblock {\emph{\JournalTitle{Journal of Land Use Science}}} \textbf{\bibinfo{volume}{17}}, \bibinfo{pages}{407--428}, \doiprefix\url{10.1080/1747423X.2022.2097455} (\bibinfo{year}{2022}).

\bibitem{ruf2015}
\bibinfo{author}{Ruf, F.}, \bibinfo{author}{Schroth, G.} \& \bibinfo{author}{Doffangui, K.}
\newblock \bibinfo{journal}{\bibinfo{title}{Climate change, cocoa migrations and deforestation in west africa: What does the past tell us about the future?}}
\newblock {\emph{\JournalTitle{Sustainability Science}}} \textbf{\bibinfo{volume}{10}}, \bibinfo{pages}{101--111}, \doiprefix\url{10.1007/s11625-014-0282-4} (\bibinfo{year}{2015}).

\bibitem{Abdulai2018}
\bibinfo{author}{Abdulai, I.} \emph{et~al.}
\newblock \bibinfo{journal}{\bibinfo{title}{Cocoa agroforestry is less resilient to sub-optimal and extreme climate than cocoa in full sun}}.
\newblock {\emph{\JournalTitle{Global Change Biology}}} \textbf{\bibinfo{volume}{24}}, \bibinfo{pages}{273--286}, \doiprefix\url{https://doi.org/10.1111/gcb.13885} (\bibinfo{year}{2018}).
\newblock \eprint{https://onlinelibrary.wiley.com/doi/pdf/10.1111/gcb.13885}.

\bibitem{schroth2016b}
\bibinfo{author}{Schroth, G.}, \bibinfo{author}{Laderach, P.}, \bibinfo{author}{Martinez-Valle, A.~I.}, \bibinfo{author}{Bunn, C.} \& \bibinfo{author}{Jassogne, L.}
\newblock \bibinfo{journal}{\bibinfo{title}{Vulnerability to climate change of cocoa in {West} {Africa}: Patterns, opportunities and limits to adaptation}}.
\newblock {\emph{\JournalTitle{Science of the Total Environment}}} \textbf{\bibinfo{volume}{556}}, \bibinfo{pages}{231--241}, \doiprefix\url{10.1016/j.scitotenv.2016.03.024} (\bibinfo{year}{2016}).

\bibitem{schroth2017}
\bibinfo{author}{Schroth, G.}, \bibinfo{author}{Läderach, P.}, \bibinfo{author}{Martinez-Valle, A.~I.} \& \bibinfo{author}{Bunn, C.}
\newblock \bibinfo{journal}{\bibinfo{title}{From site-level to regional adaptation planning for tropical commodities: cocoa in {West} {Africa}}}.
\newblock {\emph{\JournalTitle{Mitigation and Adaptation Strategies for Global Change}}} \textbf{\bibinfo{volume}{22}}, \bibinfo{pages}{903--927}, \doiprefix\url{10.1007/s11027-016-9707-y} (\bibinfo{year}{2017}).

\bibitem{niether2020}
\bibinfo{author}{Niether, W.}, \bibinfo{author}{Jacobi, J.}, \bibinfo{author}{Blaser, W.~J.}, \bibinfo{author}{Andres, C.} \& \bibinfo{author}{Armengot, L.}
\newblock \bibinfo{journal}{\bibinfo{title}{Cocoa agroforestry systems versus monocultures: a multi-dimensional meta-analysis}}.
\newblock {\emph{\JournalTitle{Environmental Research Letters}}} \textbf{\bibinfo{volume}{15}}, \bibinfo{pages}{104085}, \doiprefix\url{10.1088/1748-9326/abb053} (\bibinfo{year}{2020}).

\bibitem{bunn19}
\bibinfo{author}{Bunn, C.} \emph{et~al.}
\newblock \bibinfo{journal}{\bibinfo{title}{Recommendation domains to scale out climate change adaptation in cocoa production in {Ghana}}}.
\newblock {\emph{\JournalTitle{Climate Services}}} \textbf{\bibinfo{volume}{16}}, \bibinfo{pages}{100123}, \doiprefix\url{https://doi.org/10.1016/j.cliser.2019.100123} (\bibinfo{year}{2019}).

\bibitem{epa2007}
\bibinfo{author}{{Environmental Protection Agency}}.
\newblock \bibinfo{title}{Ghana's national definition of forest}.
\newblock \bibinfo{howpublished}{[Online]. Available from: \url{https://cdm.unfccc.int/DNA/cdf/files/2008/1706_ghana.pdf}} (\bibinfo{year}{2007}).
\newblock \bibinfo{note}{Accessed: 2024-02-01}.

\bibitem{ashiagbor2020}
\bibinfo{author}{Ashiagbor, G.} \emph{et~al.}
\newblock \bibinfo{journal}{\bibinfo{title}{Pixel-based and object-oriented approaches in segregating cocoa from forest in the {Juabeso-Bia} landscape of {Ghana}}}.
\newblock {\emph{\JournalTitle{Remote Sensing Applications: Society and Environment}}} \textbf{\bibinfo{volume}{19}}, \bibinfo{pages}{100349}, \doiprefix\url{https://doi.org/10.1016/j.rsase.2020.100349} (\bibinfo{year}{2020}).

\bibitem{chapman2020}
\bibinfo{author}{Chapman, M.} \emph{et~al.}
\newblock \bibinfo{journal}{\bibinfo{title}{Large climate mitigation potential from adding trees to agricultural lands}}.
\newblock {\emph{\JournalTitle{Global Change Biology}}} \textbf{\bibinfo{volume}{26}}, \bibinfo{pages}{4357--4365}, \doiprefix\url{10.1111/gcb.15121} (\bibinfo{year}{2020}).

\bibitem{zomer2016}
\bibinfo{author}{Zomer, R.~J.} \emph{et~al.}
\newblock \bibinfo{journal}{\bibinfo{title}{Global tree cover and biomass carbon on agricultural land: The contribution of agroforestry to global and national carbon budgets}}.
\newblock {\emph{\JournalTitle{Scientific Reports}}} \textbf{\bibinfo{volume}{6}}, \bibinfo{pages}{29987}, \doiprefix\url{10.1038/srep29987} (\bibinfo{year}{2016}).

\bibitem{nair2012}
\bibinfo{author}{Nair, P. K.~R.}
\newblock \emph{\bibinfo{title}{Climate Change Mitigation: A Low-Hanging Fruit of Agroforestry}}, vol.~\bibinfo{volume}{9} of \emph{\bibinfo{series}{Advances in Agroforestry}}, \bibinfo{type}{book section}~\bibinfo{chapter}{7}, \bibinfo{pages}{31--67} (\bibinfo{publisher}{Springer Netherlands}, \bibinfo{year}{2012}).

\bibitem{asigbaase2021}
\bibinfo{author}{Asigbaase, M.}, \bibinfo{author}{Dawoe, E.}, \bibinfo{author}{Lomax, B.~H.} \& \bibinfo{author}{Sjogersten, S.}
\newblock \bibinfo{journal}{\bibinfo{title}{Biomass and carbon stocks of organic and conventional cocoa agroforests, {Ghana}}}.
\newblock {\emph{\JournalTitle{Agriculture, Ecosystems \& Environment}}} \textbf{\bibinfo{volume}{306}}, \doiprefix\url{10.1016/j.agee.2020.107192} (\bibinfo{year}{2021}).

\bibitem{asare2018}
\bibinfo{author}{Asare, R.}, \bibinfo{author}{Markussen, B.}, \bibinfo{author}{Asare, R.}, \bibinfo{author}{Anim-Kwapong, G.} \& \bibinfo{author}{Ræbild, A.}
\newblock \bibinfo{journal}{\bibinfo{title}{On-farm cocoa yields increase with canopy cover of shade trees in two agroecological zones in ghana}}.
\newblock {\emph{\JournalTitle{Climate and Development}}} \doiprefix\url{10.1080/17565529.2018.1442805.https://doi.org/10.1080/17565529.2018.1442805} (\bibinfo{year}{2018}).

\bibitem{ma2021}
\bibinfo{author}{Ma, H.} \emph{et~al.}
\newblock \bibinfo{journal}{\bibinfo{title}{The global distribution and environmental drivers of aboveground versus belowground plant biomass}}.
\newblock {\emph{\JournalTitle{Nature Ecology \& Evolution}}} \textbf{\bibinfo{volume}{5}}, \bibinfo{pages}{1110--1122}, \doiprefix\url{10.1038/s41559-021-01485-1} (\bibinfo{year}{2021}).

\bibitem{jackson1996}
\bibinfo{author}{Jackson, R.~B.} \emph{et~al.}
\newblock \bibinfo{journal}{\bibinfo{title}{A global analysis of root distributions for terrestrial biomes}}.
\newblock {\emph{\JournalTitle{Oecologia}}} \textbf{\bibinfo{volume}{108}}, \bibinfo{pages}{389--411}, \doiprefix\url{10.1007/BF00333714} (\bibinfo{year}{1996}).

\bibitem{watson2018}
\bibinfo{author}{Watson, J. E.~M.} \emph{et~al.}
\newblock \bibinfo{journal}{\bibinfo{title}{The exceptional value of intact forest ecosystems}}.
\newblock {\emph{\JournalTitle{Nature Ecology {\&} Evolution}}} \textbf{\bibinfo{volume}{2}}, \bibinfo{pages}{599--610}, \doiprefix\url{10.1038/s41559-018-0490-x} (\bibinfo{year}{2018}).

\bibitem{lin2007}
\bibinfo{author}{Lin, B.~B.}
\newblock \bibinfo{journal}{\bibinfo{title}{Agroforestry management as an adaptive strategy against potential microclimate extremes in coffee agriculture}}.
\newblock {\emph{\JournalTitle{Agricultural and Forest Meteorology}}} \textbf{\bibinfo{volume}{144}}, \bibinfo{pages}{85--94} (\bibinfo{year}{2007}).

\bibitem{ruf2011}
\bibinfo{author}{Ruf, F.~O.}
\newblock \bibinfo{journal}{\bibinfo{title}{The myth of complex cocoa agroforests: The case of ghana}}.
\newblock {\emph{\JournalTitle{Human Ecology}}} \textbf{\bibinfo{volume}{39}}, \bibinfo{pages}{373--388}, \doiprefix\url{10.1007/s10745-011-9392-0} (\bibinfo{year}{2011}).

\bibitem{anderegg2020}
\bibinfo{author}{Anderegg, W. R.~L.} \emph{et~al.}
\newblock \bibinfo{journal}{\bibinfo{title}{Climate-driven risks to the climate mitigation potential of forests}}.
\newblock {\emph{\JournalTitle{Science}}} \textbf{\bibinfo{volume}{368}}, \bibinfo{pages}{eaaz7005}, \doiprefix\url{doi:10.1126/science.aaz7005} (\bibinfo{year}{2020}).

\bibitem{unruh2008}
\bibinfo{author}{Unruh, J.~D.}
\newblock \bibinfo{journal}{\bibinfo{title}{Carbon sequestration in africa: The land tenure problem}}.
\newblock {\emph{\JournalTitle{Global Environmental Change}}} \textbf{\bibinfo{volume}{18}}, \bibinfo{pages}{700--707}, \doiprefix\url{https://doi.org/10.1016/j.gloenvcha.2008.07.008} (\bibinfo{year}{2008}).

\bibitem{reichstein2013}
\bibinfo{author}{Reichstein, M.} \emph{et~al.}
\newblock \bibinfo{journal}{\bibinfo{title}{Climate extremes and the carbon cycle}}.
\newblock {\emph{\JournalTitle{Nature}}} \textbf{\bibinfo{volume}{500}}, \bibinfo{pages}{287--295}, \doiprefix\url{10.1038/nature12350} (\bibinfo{year}{2013}).

\bibitem{blaser_2021}
\bibinfo{author}{Blaser-Hart, W.~J.} \emph{et~al.}
\newblock \bibinfo{journal}{\bibinfo{title}{The effectiveness of cocoa agroforests depends on shade-tree canopy height}}.
\newblock {\emph{\JournalTitle{Agriculture, Ecosystems \& Environment}}} \textbf{\bibinfo{volume}{322}}, \bibinfo{pages}{107676}, \doiprefix\url{https://doi.org/10.1016/j.agee.2021.107676} (\bibinfo{year}{2021}).

\bibitem{somarriba2024}
\bibinfo{author}{Somarriba, E.}, \bibinfo{author}{Saj, S.}, \bibinfo{author}{Orozco-Aguilar, L.}, \bibinfo{author}{Somarriba, A.} \& \bibinfo{author}{Rapidel, B.}
\newblock \bibinfo{journal}{\bibinfo{title}{Shade canopy density variables in cocoa and coffee agroforestry systems}}.
\newblock {\emph{\JournalTitle{Agroforestry Systems}}} \textbf{\bibinfo{volume}{98}}, \bibinfo{pages}{585--601}, \doiprefix\url{10.1007/s10457-023-00931-2} (\bibinfo{year}{2024}).

\bibitem{zuidema2005}
\bibinfo{author}{Zuidema, P.~A.}, \bibinfo{author}{Leffelaar, P.~A.}, \bibinfo{author}{Gerritsma, W.}, \bibinfo{author}{Mommer, L.} \& \bibinfo{author}{Anten, N.~P.}
\newblock \bibinfo{journal}{\bibinfo{title}{A physiological production model for cocoa (theobroma cacao): model presentation, validation and application}}.
\newblock {\emph{\JournalTitle{Agricultural Systems}}} \textbf{\bibinfo{volume}{84}}, \bibinfo{pages}{195--225}, \doiprefix\url{https://doi.org/10.1016/j.agsy.2004.06.015} (\bibinfo{year}{2005}).

\bibitem{hainmueller2011sustainable}
\bibinfo{author}{Hainmueller, J.}, \bibinfo{author}{Hiscox, M.} \& \bibinfo{author}{Tampe, M.}
\newblock \bibinfo{journal}{\bibinfo{title}{Sustainable development for cocoa farmers in {Ghana}}}.
\newblock {\emph{\JournalTitle{{International Growth Centre, London School of Economics}}}}  (\bibinfo{year}{2011}).

\bibitem{bymolt2018demystifying}
\bibinfo{author}{Bymolt, R.}, \bibinfo{author}{Laven, A.} \& \bibinfo{author}{Tyzler, M.}
\newblock \bibinfo{journal}{\bibinfo{title}{Demystifying the cocoa sector in {Ghana} and {C{\^o}te d'Ivoire}}}.
\newblock {\emph{\JournalTitle{The Royal Tropical Institute (KIT), Amsterdam}}}  (\bibinfo{year}{2018}).

\bibitem{franzen2007}
\bibinfo{author}{Franzen, M.} \& \bibinfo{author}{Borgerhoff~Mulder, M.}
\newblock \bibinfo{journal}{\bibinfo{title}{Ecological, economic and social perspectives on cocoa production worldwide}}.
\newblock {\emph{\JournalTitle{Biodiversity and Conservation}}} \textbf{\bibinfo{volume}{16}}, \bibinfo{pages}{3835--3849}, \doiprefix\url{10.1007/s10531-007-9183-5} (\bibinfo{year}{2007}).

\bibitem{friedman2001greedy}
\bibinfo{author}{Friedman, J.~H.}
\newblock \bibinfo{journal}{\bibinfo{title}{Greedy function approximation: a gradient boosting machine}}.
\newblock {\emph{\JournalTitle{Annals of Statistics}}} \bibinfo{pages}{1189--1232} (\bibinfo{year}{2001}).

\bibitem{huber64}
\bibinfo{author}{Huber, P.~J.}
\newblock \bibinfo{journal}{\bibinfo{title}{Robust estimation of a location parameter}}.
\newblock {\emph{\JournalTitle{Annals of Statistics}}} \textbf{\bibinfo{volume}{53}}, \bibinfo{pages}{73–101} (\bibinfo{year}{1964}).

\bibitem{abderazak96}
\bibinfo{author}{Abderrazak, B.}, \bibinfo{author}{Morin, D.}, \bibinfo{author}{Bonn, F.} \& \bibinfo{author}{Huete, A.}
\newblock \bibinfo{journal}{\bibinfo{title}{A review of vegetation indices}}.
\newblock {\emph{\JournalTitle{Remote Sensing Reviews}}} \textbf{\bibinfo{volume}{13}}, \bibinfo{pages}{95--120}, \doiprefix\url{10.1080/02757259509532298} (\bibinfo{year}{1996}).

\bibitem{Moraiti2024}
\bibinfo{author}{Moraiti, N.}, \bibinfo{author}{Mullissa, A.}, \bibinfo{author}{Rahn, E.}, \bibinfo{author}{Sassen, M.} \& \bibinfo{author}{Reiche, J.}
\newblock \bibinfo{journal}{\bibinfo{title}{Critical assessment of cocoa classification with limited reference data: A study in côte d’ivoire and ghana using sentinel-2 and random forest model}}.
\newblock {\emph{\JournalTitle{Remote Sensing}}} \textbf{\bibinfo{volume}{16}}, \doiprefix\url{10.3390/rs16030598} (\bibinfo{year}{2024}).

\bibitem{QGIS}
\bibinfo{author}{{QGIS Development Team}}.
\newblock \emph{\bibinfo{title}{QGIS Geographic Information System}}.
\newblock \bibinfo{organization}{Open Source Geospatial Foundation Project} (\bibinfo{year}{2023}).
\newblock \bibinfo{note}{Version 3.34.12}.

\bibitem{duncanson2022aboveground}
\bibinfo{author}{Duncanson, L.} \emph{et~al.}
\newblock \bibinfo{journal}{\bibinfo{title}{Aboveground biomass density models for {NASA}’s global ecosystem dynamics investigation ({GEDI}) lidar mission}}.
\newblock {\emph{\JournalTitle{Remote Sensing of Environment}}} \textbf{\bibinfo{volume}{270}}, \bibinfo{pages}{112845} (\bibinfo{year}{2022}).

\bibitem{saarela2018generalized}
\bibinfo{author}{Saarela, S.} \emph{et~al.}
\newblock \bibinfo{journal}{\bibinfo{title}{Generalized hierarchical model-based estimation for aboveground biomass assessment using {GEDI} and {Landsat} data}}.
\newblock {\emph{\JournalTitle{Remote Sensing}}} \textbf{\bibinfo{volume}{10}}, \bibinfo{pages}{1832} (\bibinfo{year}{2018}).

\bibitem{dubayah2022gedi}
\bibinfo{author}{Dubayah, R.} \emph{et~al.}
\newblock \bibinfo{title}{{GEDI} {L4A} footprint level aboveground biomass density, version 2.1}.
\newblock \bibinfo{type}{Tech. Rep.}, \bibinfo{institution}{{ORNL} {DAAC}, {Oak Ridge}, {Tennessee}, {USA}} (\bibinfo{year}{2022}).

\bibitem{lanfranchi22}
\bibinfo{author}{Lanfranchi, C.}
\newblock \bibinfo{title}{Global biomass mapping and uncertainty estimation from {GEDI} {LIDAR} data using bayesian deep learning}.
\newblock \bibinfo{type}{Tech. Rep.}, \bibinfo{institution}{MSc thesis in Data Science, ETH Zurich} (\bibinfo{year}{2022}).

\bibitem{lefsky05}
\bibinfo{author}{Lefsky, M.~A.} \emph{et~al.}
\newblock \bibinfo{journal}{\bibinfo{title}{Estimates of forest canopy height and aboveground biomass using {ICESat}}}.
\newblock {\emph{\JournalTitle{Geophysical Research Letters}}} \textbf{\bibinfo{volume}{32}}, \doiprefix\url{https://doi.org/10.1029/2005GL023971} (\bibinfo{year}{2005}).

\bibitem{asner12}
\bibinfo{author}{Asner, G.~P.} \emph{et~al.}
\newblock \bibinfo{journal}{\bibinfo{title}{High-resolution mapping of forest carbon stocks in the {Colombian} {Amazon}}}.
\newblock {\emph{\JournalTitle{Biogeosciences}}} \textbf{\bibinfo{volume}{9}}, \bibinfo{pages}{2683--2696}, \doiprefix\url{10.5194/bg-9-2683-2012} (\bibinfo{year}{2012}).

\bibitem{KingmaB14}
\bibinfo{author}{Kingma, D.~P.} \& \bibinfo{author}{Ba, J.}
\newblock \bibinfo{title}{Adam: {A} method for stochastic optimization}.
\newblock In \bibinfo{editor}{Bengio, Y.} \& \bibinfo{editor}{LeCun, Y.} (eds.) \emph{\bibinfo{booktitle}{Proceedings of the International Conference on Learning Representations}} (\bibinfo{year}{2015}).

\bibitem{vancutsem2021}
\bibinfo{author}{Vancutsem, C.} \emph{et~al.}
\newblock \bibinfo{journal}{\bibinfo{title}{Long-term (1990–2019) monitoring of forest cover changes in the humid tropics}}.
\newblock {\emph{\JournalTitle{Science Advances}}} \textbf{\bibinfo{volume}{7}}, \bibinfo{pages}{eabe1603}, \doiprefix\url{doi:10.1126/sciadv.abe1603} (\bibinfo{year}{2021}).

\bibitem{martin2011}
\bibinfo{author}{Martin, A.~R.} \& \bibinfo{author}{Thomas, S.~C.}
\newblock \bibinfo{journal}{\bibinfo{title}{A reassessment of carbon content in tropical trees}}.
\newblock {\emph{\JournalTitle{PLOS ONE}}} \textbf{\bibinfo{volume}{6}}, \bibinfo{pages}{e23533}, \doiprefix\url{10.1371/journal.pone.0023533} (\bibinfo{year}{2011}).

\bibitem{blaser-hart2025cocoa}
\bibinfo{author}{Blaser-Hart, W.~J.}
\newblock \bibinfo{title}{Cocoa agroforestry shade and biomass data for ground truthing and climate mitigation analysis (ghana and côte d’ivoire, 2021–2022)}, \doiprefix\url{10.48610/dda018c} (\bibinfo{year}{2025}).
\newblock \bibinfo{note}{Dataset}.

\end{thebibliography}
\defaultbibliography{main}



\vspace{1.8em}
\section*{Extended data figures}
\setcounter{figure}{0}
\renewcommand{\figurename}{Fig.}
\renewcommand{\thefigure}{Extended Data \arabic{figure}}
\renewcommand{\thetable}{Extended Data \arabic{table}}

\begin{figure}[hb!]
    \centering
    \includegraphics[width=1\textwidth]{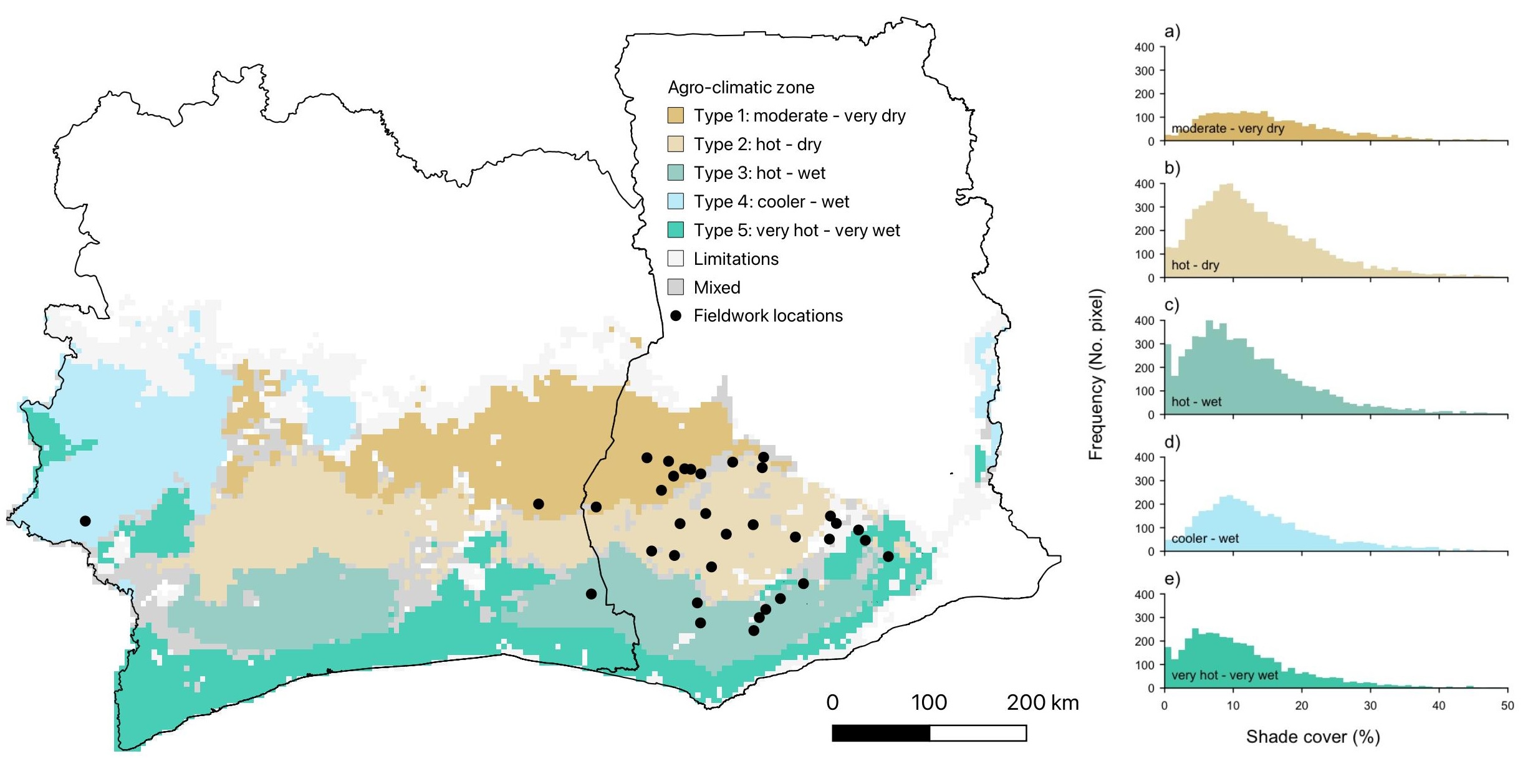}
    \caption{Agro-climatic classifications and shade levels across cocoa-growing regions of Côte d'Ivoire and Ghana. Different colors represent various climatic types determined from climatic data spanning 1990-2020. Areas in dark grey (mixed) show classification uncertainty, while light grey areas (limitations) are likely unsuitable for cocoa cultivation but with high uncertainty, and white areas are unsuitable for cocoa growth. Black dots indicate cocoa communities visited during fieldwork for ground truth data collection. Sub-panels (a-e) display the shade levels in each agro-climatic zone based on our map of shade-tree cover (Fig.~\ref{fig:shade_map}). Base map boundaries were derived from the Global Administrative Areas database (GADM v4.1, \href{https://gadm.org/}{https://gadm.org}).}
    \label{fig:aez}
\end{figure}

\begin{figure}[ht]
    \centering
    \includegraphics[width=1\textwidth]{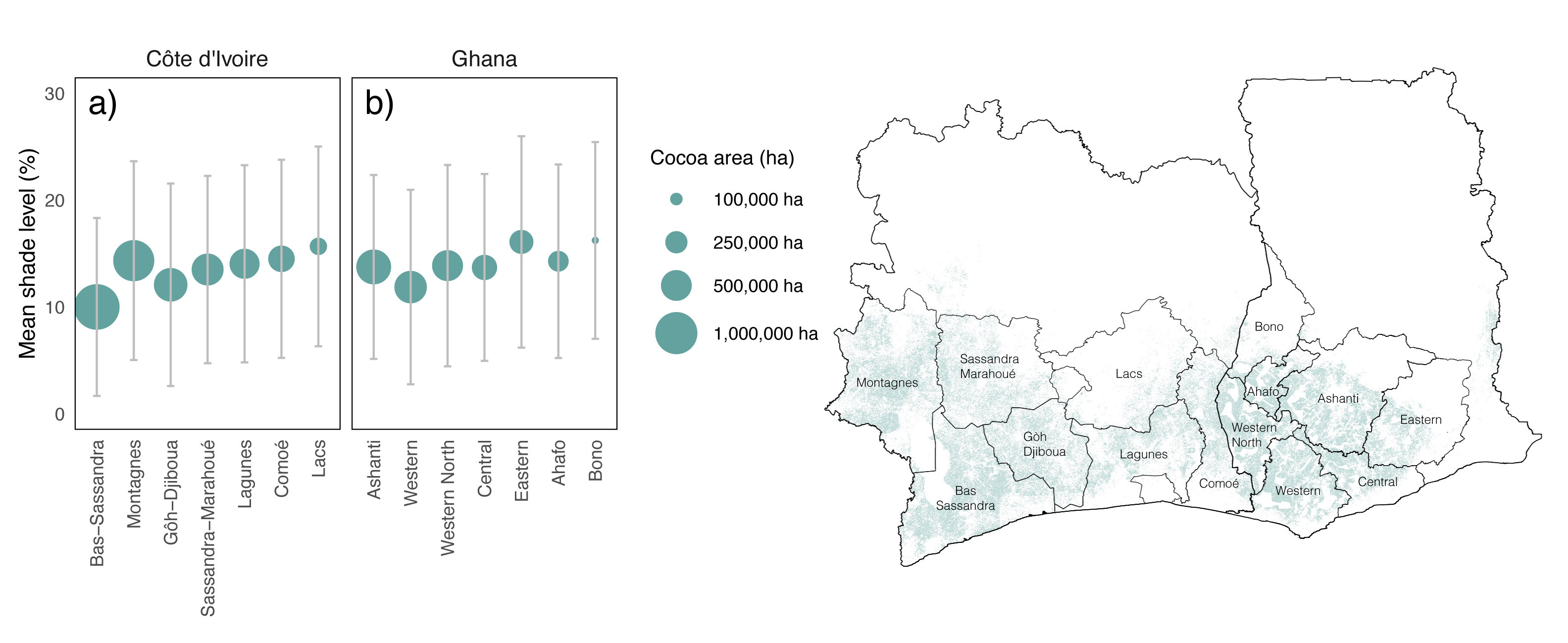}
    \caption{Shade levels in a) cocoa-growing districts of Côte d'Ivoire and b) cocoa-growing regions of Ghana. The maps display the distribution of administrative districts/regions within cocoa-growing areas, shaded in turquoise. Mean shade levels (\%) are shown for each district/region, with error bars representing $\pm$ one standard deviation across all pixel-level values. The unit of analysis is individual 10$\times$10~m pixels, with a total of $n = 679,593,915$ pixels included across all district/region. Sample sizes (n) per region correspond to the number of pixels within each administrative unit and vary with cocoa-growing area. Bubble sizes corresponds to the total cocoa-growing area in each region, ranging from 78,382 to 1,190,000 hectares. Only administrative regions/districts comprising more than 2\% of the total cocoa-growing area in each country were included in the plot. All data represent biological replicates based on independent spatial observations. Cocoa-growing areas are from Kalischek et al. (2023)\cite{kalischek23natfood}. Base map boundaries were derived from the Global Administrative Areas database (GADM v4.1, \url{https://gadm.org/}).}
    \label{fig:regions_shade}
\end{figure}

\begin{figure}[th]
    \centering
    \includegraphics[width=0.9\textwidth]{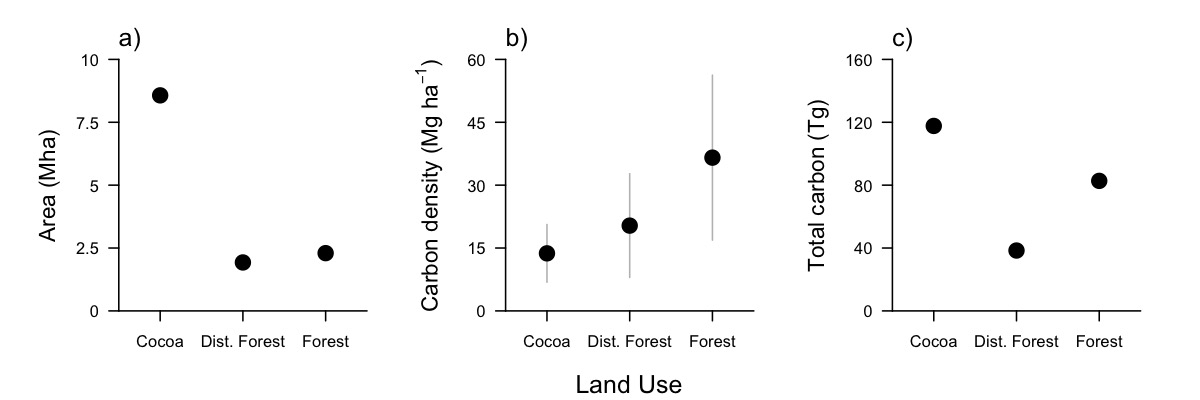}
    \caption{Area, carbon densities, and carbon stocks of cocoa-growing areas, disturbed forests (dist. forest), and undisturbed forests (forest) across Ghana and Côte d’Ivoire. The values in (a) represent the total mapped extent of each land use class, for which there are no error estimates. The values in (b) are means $\pm$ standard deviations, because we are interested in variation among land use class in carbon density. The values in (c) are total carbon $\pm$ 95\% confidence intervals, reflecting uncertainty in the estimates of total carbon stocks. All values were calculated at the resolution of the AGBD map (50$\times$50~m). The unit of analysis is individual pixels. Sample sizes ($n$) correspond to the number of pixels per land cover class: $n = 34,284,584$ for cocoa, $n = 7,704,695$ for disturbed forest, and $n = 9,185,882$ for undisturbed forest. Due to these large sample sizes for pixel-level estimates of total carbon, the confidence intervals in panel (c) are too small to be visually distinguishable. Reported 95\% confidence intervals are: cocoa = $\pm 1.2 \times 10^{-5}$ million tonnes C; disturbed forest = $\pm 4.1 \times 10^{-6}$ million tonnes C; undisturbed forest = $\pm 2.5 \times 10^{-6}$ million tonnes C. All data represent biological replicates in the form of independent spatial observations.}
    \label{LastPagePart1}
\label{fig:landuse_carbon_stocks}
\end{figure}




\clearpage

\appendix

\setcounter{page}{1}
\setcounter{figure}{0}
\renewcommand{\thefigure}{S\arabic{figure}}
\renewcommand{\thetable}{S\arabic{table}}

\part*{Supplementary information}

\fancyfoot{}  
\fancyfoot[R]{\small\sffamily\bfseries\thepage/\pageref{LastPagePart2}}

\begin{bibunit}

\section*{Supplementary methods}
\subsection*{Agro-climatic zones in Ghana and Côte d’Ivoire.}
Agro-climatic zoning (ACZ) is widely used to delineate comparable environmental regions in order to guide targeted interventions. No existing agro-climatic classification consistently covered both Ghana and Côte d’Ivoire in a crop-relevant way, so we developed a new scheme tailored to cocoa production systems across the region. To delineate agro-climatic zones, we grouped cocoa-growing areas with similar climate conditions using the Random Forest (RF) classifier \cite{breiman2001}, following the approach of Kamath et al. \cite{Kamath2024}. However, in contrast to Kamath et al. \cite{Kamath2024}, we sampled cocoa occurrences from Kalischek et al.\cite{kalischek23natfood} and clustered them using more recent climate data to identify distinct agro-climatic types. The RF classifier was then trained and extrapolated to generate maps of the agro-climatic zones.

\textit{Occurrence data.} We obtained cocoa occurrence data from high-resolution maps of cocoa-growing areas in Côte d’Ivoire and Ghana \cite{kalischek23natfood}. This data consisted of cocoa plots identified on Copernicus Sentinel-2 optical satellite imagery using a deep learning framework and <100.000 GPS polygon mapped cocoa farms. We first geographically subsampled at 2.5’ from this layer. We obtained <9000 unique cocoa locations, from which we drew a 10\% random sample as reference distribution.

\textit{Climate analysis.} We constructed a monthly climatology using temperature data from CHIRTS-ERA5 \cite{funk2019} and precipitation data from CHIRPS v2.0 \cite{funk2014}, spanning January 1990 to December 2020. Both datasets were originally available at daily resolution and 0.05° spatial resolution. Solar radiation inputs were obtained from the Global PET and Aridity Index database \cite{trabucco2018} and used to calculate potential evapotranspiration (PET) via the modified Hargreaves method \cite{Allan1998}.

To characterize climatic suitability for cocoa, we followed the approach of Kamath et al. \cite{Kamath2024}, deriving 33 bioclimatic variables: the standard 19 variables from WorldClim, along with cocoa-relevant metrics such as the number of consecutive dry months, arid periods (where precipitation is less than PET), total dry-season water deficit, growing season temperatures, and dry-season maximum temperatures. To reduce multicollinearity, we applied variance inflation factor (VIF) analysis, resulting in a final subset of 13 variables used to model agro-climatic zones (see Supplementary Fig.~\ref{fig:bioc_boxplot} for the selected variables).

\textit{Definition of agro-climate types.} Following the methods outlined in Kamath et al. \cite{Kamath2024}, we used the Random Forest (RF) classifier \cite{breiman2001} in two main steps: (1) identifying agro-climatic clusters in cocoa-growing regions, and (2), classifying current climate conditions into these types. RF was chosen for its robustness in classification tasks and low risk of overfitting. We implemented the model using the \textit{randomForest} package \cite{liaw2002} in R \cite{R2021}.

In the first step, we identified agro-climatic clusters by calculating dissimilarities at cocoa occurrence locations using 13 selected climate variables \cite{shi2006}. We then grouped these locations using Ward hierarchical clustering, selecting the number of clusters based on visual inspection of the resulting dendrogram.

In the second step, we classified current climate conditions into the identified agro-climatic clusters. To reduce bias and exclude climatically unrealistic areas, we combined cocoa occurrence points with a background sample (1:1 ratio) drawn randomly from cocoa-producing countries \cite{VanDerWal2009}. Each RF model was configured with 100 forests and 1000 trees, and repeated 25 times to ensure stability. We split the data into 80\% for training and 20\% for evaluation. The final classification maps represent the modal prediction across all trees, with “mixed” labels assigned to ambiguous locations and “limitations” for areas with partial suitability.

\textit{Results.}We identified five distinct agro-climatic types across Ghana and Côte d'Ivoire, each showing unique patterns of cocoa suitability. To interpret and describe these types, we examined the confidence intervals of each climate variable across the RF-defined clusters (Supplementary Fig.~\ref{fig:bioc_boxplot}). Based on their distinguishing climate features, we described the five agro-climatic types as follows:

\begin{enumerate}
    \item \textbf{Type 1: Moderate – very dry:} Intermediate values for most temperature variables (Bio 3, 4, 8), but low values for precipitation during the driest month (Bio 14) and high number of consecutive months with less than 100 mm of precipitation (Bio 20).
    \item \textbf{Type 2: Hot \& dry:} Relatively high temperature values (Bio 8 and 11), high precipitation seasonality (Bio 15), long dry season (Bio 20).
    \item \textbf{Type 3: Hot \& wet:} Relatively high temperature values (Bio 8 and 11), high precipitation during the driest month (Bio 14), lowest number of consecutive dry months (Bio 20).
    \item \textbf{Type 4: Cooler} \& wet: Low mean temperature of the wettest quarter (Bio 8), high precipitation of the coldest quarter (Bio 19), relatively high number of consecutive dry months (Bio 20), high potential evapotranspiration (Bio 25).
    \item \textbf{Type 5: Very hot \& very wet:} High temperature seasonality and values (Bio 4, 8, and 11), low number of consecutive dry months (Bio 20).
\end{enumerate}
These five agro-climatic types were mapped as five agro-climatic zones for Ivory Coast and Ghana, along with three additional classifications: "unsuitable" zones, which are unlikely to support cocoa due to climatic constraints,  "mixed" zones which could not be clearly categorized into one of the agro-climatic zones and "limitations" zones which were classified unsuitable but were found to contain cocoa occurrences (Extended Data Figure \ref{fig:aez}). This mapping provides a comprehensive view of the suitability of different areas for cocoa cultivation, considering the varying constellations of climatic attributes. Briefly, the Type 1 – moderate- very dry zone spanned the Northern cocoa zone from Yamoussoukro (CdI) to Ahafo (Ghana). Type 2 – hot \& dry climate can be found South of Type 1 and North of Type 3 from Goh (CdI) to Ashanti (Ghana). The central cocoa zones Soubre and Divo in CdI, as well as Western Region in Ghana were characterized by Type 3 - hot \& wet climate. Type 4 – cooler \& wet – climate can be found in the Montagnes area (CdI) and the Ghanaian boarder with Togo, which also has some higher elevation area. Type 5 – very hot \& very wet climate mostly covered coastal areas along the Gulf of Guinea (Extended Data Figure \ref{fig:aez}). 

\subsection*{Sampling design for shade-tree ground-truth data}
To distribute our ground truth sampling effort in Ghana, we first randomly selected eight administrative districts within each of four agro-climatic zones. The four agro-climatic zones were defined earlier by Bunn et al.~\cite{bunn19}, using a random forest machine learning algorithm to classify cocoa growing regions in Ghana according to climate and soil characteristics. Within each administrative district we randomly chose either one or two cocoa-growing communities, with the constraint that the selected communities have at least 25 recently mapped cocoa farms. We then randomly picked between 16 and 39 farms per community, ensuring that farms were distributed relatively evenly across the community, and that each individual farm was managed by a different farmer where possible. 

Ground-truthing in Côte d’Ivoire was done in collaboration with a sampling program initiated by the Sustainable Cocoa Initiative Support Programme (SCISP) and the Sustainable Agicultural Supply Chains Initiative (SASI) of the German Development Agency (GIZ). These farms were distributed in the Sud Comoé and Indénié-Djuablin administrative regions in the east and Cavally region in the west of Côte d’Ivoire. Subsequent to this, we extended the agro-climatic zoning methods of Bunn et al. ~\cite{bunn19} to include Côte d’Ivoire (Supplementary Methods, Extended Data Fig.~\ref{fig:aez}), which indicated that these 60 farms were distributed across two of the four original cocoa agro-climatic zones identified in Ghana, and an additional cool and wet zone in the west of Côte d’Ivoire that was not originally identified in Ghana, where it only occurs in the Volta region (Extended Data Fig.~\ref{fig:aez}). We calculate all results with respect to the updated zones across both countries.

To identify the additional 69 farms with higher cover, we relied on farmer-reported shade-tree densities combined with interactive visual interpretation of Google Earth images. These additional high-cover farms were distributed in 30 cocoa-growing communities across four administrative districts in two agro-climatic zones. 

\subsection*{Generating digital surface and terrain models}
We used DroneDeploy software (DroneDeploy Inc., Version 2.236.0) to generate a digital surface model (DSM; top of canopy) and a digital terrain model (DTM) and subtracted the two to obtain a canopy height map for each farm. The DSM is generated first using the drone imagery, which captures all visible surface features, including buildings, vegetation, and other structures. The point cloud data from the drone imagery is also analyzed to detect and classify ground points, of which there tended to be many identified for each farm. To create the DTM, the algorithm filters out all non-ground elements, isolating ground points to represent the bare-earth surface. In areas where direct ground visibility is limited, interpolation techniques are applied to estimate terrain elevation based on surrounding data. The final result is a continuous surface, though some degree of approximation is involved depending on terrain complexity and data density. We note that individual farms are relatively flat, which favors more accurate terrain interpolation based on available ground points, and also limits the chance of systematic biases in terrain estimation that could systematically bias our canopy height estimates within or between farms. Finally, our threshold height for distinguishing cocoa from shade trees (see below) is conservative, such that our methods are likely to be robust to any (likely small in both magnitude and spatial extent) positive or negative mis-estimates of terrain elevation within farms.

\subsection*{Threshold height calculations}
\begin{figure}
    \centering
    \includegraphics[width=0.6\textwidth]{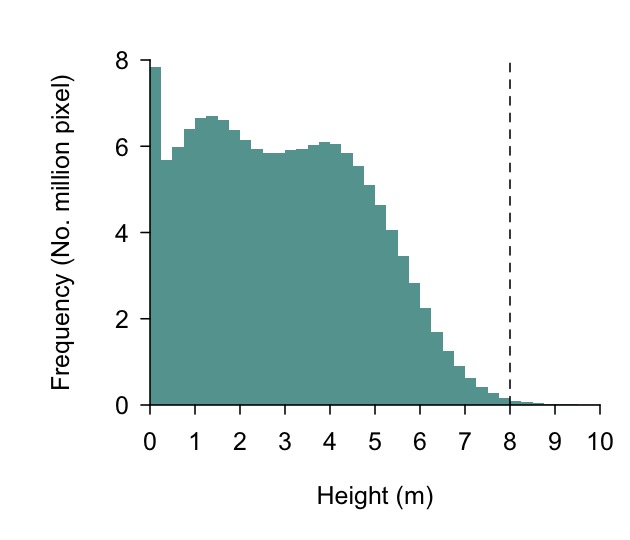}
    \caption{Distribution of heights for cocoa trees in ground truth data. Cocoa tree heights were derived from 6,800 manually-labeled polygons across small sections of cocoa monocultures, using drone-derived canopy height maps. These polygons covered 118 hectares and encompassed 230 cocoa farms throughout the study region. An 8-meter height cut-off (indicated by the dashed vertical line) was used to distinguish cocoa trees from shade trees. Notably, 99.7\% of all cocoa tree height data points fall below this threshold. \textit{N} = 149,913,724 individual pixels within the labeled cocoa polygons. No statistical tests were performed; data are shown to characterize the distribution of cocoa canopy heights}
    \label{fig:height_cutoff}
\end{figure}

To determine a suitable height threshold to distinguish between shade trees and cocoa trees (which tend to be relatively short), we manually labelled 6800 polygons delimiting small sections of cocoa monocultures (total area 118 ha) in 230 cocoa farms distributed across the study area. These monoculture areas were identified using visual interpretation of orthophotos derived from the drone images. For all labelled polygons we then extracted the height of cocoa trees from the canopy height map. This analysis showed that 99.7\% of all cocoa trees remained below a height of eight meters (Supplementary Fig.~\ref{fig:height_cutoff}), while previous work shows that heights of shade trees tend to be higher than eight meters \cite{blaser_2021}. We therefore used a threshold height of eight meters; vegetation above this height could be confidently attributed to shade trees rather than cocoa trees. 

\subsection*{Shade-cover model performance}
On our hold-out test set of 259 farms, we achieve a mean absolute error of 5.77 percentage points (pp.), a root mean squared error of 8.22 pp., and a mean (bias) error of 1.63 pp. We visualize a residual analysis for different reference shade cover values in Supplementary Fig. \ref{fig:agbd_comp}a).

\subsection*{Comparison to existing AGBD estimates}
We compare our biomass map with the European Space Agency's (ESA) Climate Change Initiative (CCI) state-of-the-art Biomass map for 2018 (\href{https://climate.esa.int/en/projects/biomass/}{ESA CCI map}). The ESA CCI map is intended to provide global coverage of AGBD estimates with a ground sampling distance of $100\times100$ meters (i.e., four times less pixels per area than our map), with the aim of monitoring global biomass change in the context of climate change. The ESA CCI map is created bi-annually based on multi-modal input data like spaceborne SAR (newer versions also include GEDI) and processed using a proprietary algorithm. Because we are lacking an independent, large-scale ground truth dataset of AGBD collected in the field across our region of interest, we compare both maps to 44,200 GEDI footprints that fall in our hold-out test areas (i.e., unseen by our model during training and validation) and contain cocoa according to Kalischek et al.\cite{kalischek23natfood}. One should keep in mind though that our map has been specifically fine-tuned to our area of interest (as opposed to the ESA CCI map, which is a global product) and it is using GEDI footprint-level AGBD data for training, too. Unknown global biases that could be present in the GEDI AGBD footprint-level product would thus most likely impact our map but not show up in the comparison to the hold-out GEDI footprints here, if those would exhibit the same kind of possible bias. This comparison shall therefore not judge the quality of the ESA CCI map but rather provide some rough notion about our map's properties. Overall, our map achieves (all values in tonnes ha\(^{-1}\)) a mean absolute error (MAE) of 36.2 (vs. ESA's 56.9), a root mean squared error (RMSE) of 63.2 (vs. ESA's 87.9) and a mean error (bias) of 6.5 (vs. ESA's 33.11). This shows that, in cocoa areas, our map does not only exhibit lower errors than the comparison map, but is also less biased towards lower AGBD values. In Supplementary Fig.~\ref{fig:agbd_comp}b, we additionally show an analysis of the residuals between the two maps and the GEDI reference footprints, to better understand the biases of both maps over different AGBD intervals.

\subsection*{Bayesian regression methods for quantifying relationship between cover and biomass}
We quantified the relationship between shade-tree cover and total biomass with Bayesian linear regression, using our pixel-level estimates of shade-tree cover as the predictor variable, and our pixel-level estimates of total aboveground biomass as the response variable. Only pixels in cocoa growing areas with shade levels up to 40\% cover were used for the analysis. That set is representative for the range of plausible agroforestry scenarios given current recommendations, and includes 99.89\% of all data points. In addition to shade-tree cover, we included a second-order polynomial term (i.e., the predictor "shade-tree-cover$^{2}$") to account for potential curvilinearity in the relationship between cover and biomass. We used a Gaussian likelihood function for the response variable, and for each model coefficient we used improper flat priors (a uniform distribution over the real numbers), a common uninformative prior in Bayesian analyses. The model was fitted using Stan (\url{https://mc-stan.org}) together with the R package \textit{brms}\cite{bürkner2017}. It was trained on a 10\% random sample of the filtered dataset and saved as an \texttt{.rds} object to support reproducibility and future analysis. The model was fitted using four Markov chain Monte Carlo (MCMC) chains, each with 2000 iterations, with a warm-up period of 1000 iterations. This gave a total of 4000 samples. Trace plots were visually assessed to ensure model convergence, and we also checked that the scale reduction factor ($\hat{R}$) was approximately equal to one. We note that we did attempt to fit an error-in-variables model to account for measurement uncertainty in the predictor, but this model consistently failed to converge, so we proceeded with the simpler Bayesian regression model as described above. With respect to predicting carbon-sequestration potential the simpler model should be more conservative as, all else being equal, it will tend to underestimate the strength of the relationship between cover and biomass.

\subsection*{Additional carbon-sequestration calculations}
To estimate annual levels of carbon-sequestration potential for the 30\% shade-tree cover target, we assume linear rates of implementation (planting) and tree growth over a 30 year time-horizon. This is consistent with a time-to-maturity of shade-trees of $\approx$20-30 years. When distributed across the area that currently has <30\% cover (6.64 million hectares), these calculations result in a carbon sequestration rate of 1.635 tonnes \ce{CO2}e ha\(^{-1}\) year\(^{-1}\). Given that estimates of carbon accumulation rates in cocoa agroforestry systems range between~ 1.3 - 11 tonnes \ce{CO2}e ha\(^{-1}\)year\(^{-1}\)\cite{nair2009,somarriba2013}, this rate is conservative, as is also appropriate given our focus on sequestration from additional shade trees alone. We compared our estimated annual sequestration rate to estimates of the total greenhouse gas emissions across both Ghana and Côte d'Ivoire. We used estimates of these countries' total greenhouse gas emissions from each country's most recent National Greenhouse Gas Inventory submitted to the UNFCC. We used the 2019 estimates, as this was the most recent year for which both countries provided data in their respective reports. 

To estimate annual emissions from cocoa production in these regions, we used estimates of greenhouse gas emissions generated per kg of cocoa (to the farm gate) for both Ghana and Côte d'Ivoire. These estimates were provided by Quantis (\url{https://quantis.com/}) using data from the World Food Life-Cycle-Assessment Database and its extensions \cite{Quantis2024}. For each country we used two different estimates in our calculations, either excluding or including the contribution of life-cycle-assessment-based, statistical land-use change (sLUC) to emissions. For the cocoa sector in West Africa, sLUC predominantly captures the emissions generated as a consequence of deforestation. For Ghana, these estimates were 2 kg and 4 kg \ce{CO2}e / kg cocoa, excluding and including sLUC, respectively. For Côte d'Ivoire, these estimates were 2 kg and 30 kg \ce{CO2}e / kg cocoa, excluding and including sLUC, respectively. The differences between the values emphasize the large contribution of land-use change to total emissions, reflecting high rates of deforestation as areas under cocoa production expanded. To get estimates of total \ce{CO2}e generated in the production of cocoa, we multiplied the per kg \ce{CO2}e estimates for each country by the average annual total cocoa production in each country between 2017 and 2021 \cite{FAO22}. We then calculated the percentage of the annual emissions that would be counterbalanced by the additional annual sequestration that would be allowed by meeting the 30\% shade-tree cover target. 

\clearpage

\section*{Supplementary figures}

\begin{figure}[!h]
    \centering
    \includegraphics[width=0.495\textwidth]{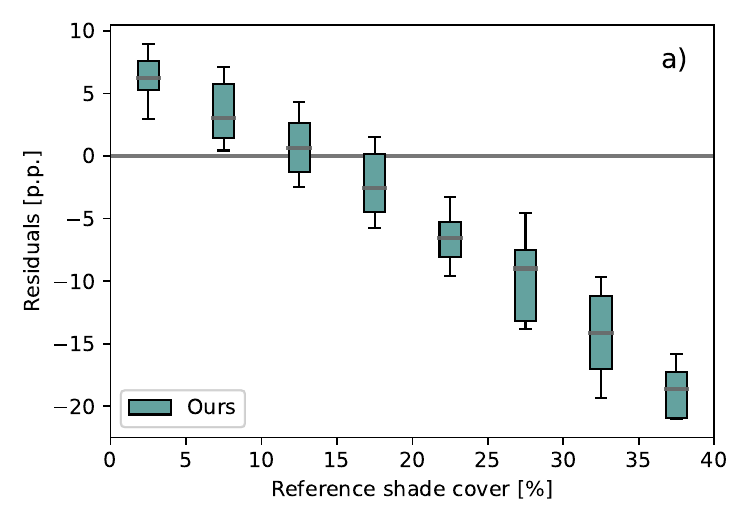}
    \includegraphics[width=0.495\textwidth]{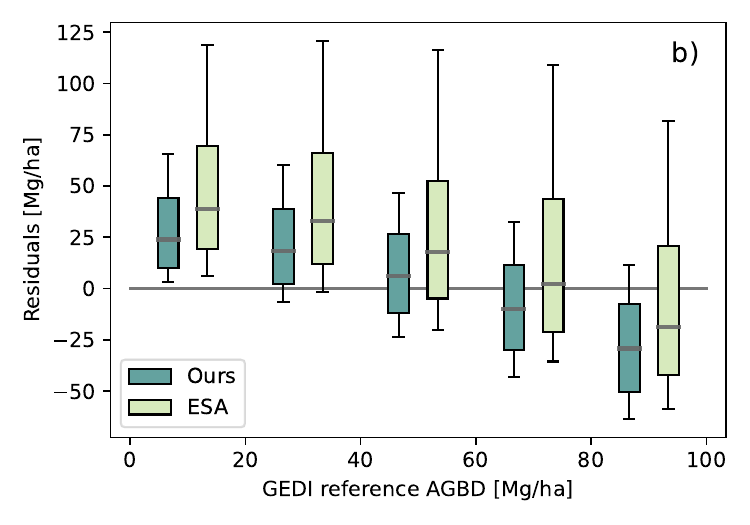}
    \caption{Residual analysis for a) shade-tree cover and b) aboveground biomass density (AGBD). a) Residual analysis of shade-tree cover intervals (each spanning 5\%) from our map, compared against ground truth field measurements. b) Residual analysis of AGBD intervals (each spanning 15 tonnes ha\(^{-1}\)) for our map and the \href{https://climate.esa.int/en/projects/biomass/}{European Space Agency's (ESA) Climate Change Initiative (CCI) Biomass map}, referenced against GEDI ground truth data. The boxplots display the median, interquartile range, and whiskers extending from the 10\(^{th}\) to 90\(^{th}\) percentiles of the residuals. Negative values indicate underestimation relative to the reference. Our map generally exhibits lower biases compared to the ESA map, except at very high AGBD values. Panel a: $n = 259$ cocoa farms in the hold-out test set. Panel b: $n = 44,220$ GEDI footprints in the hold-out test set. No statistical tests were performed; residuals are shown to assess model accuracy.}
    \label{fig:agbd_comp}
\end{figure}

\begin{figure}[!hb]
    \centering
    \includegraphics[width=0.55\textwidth]{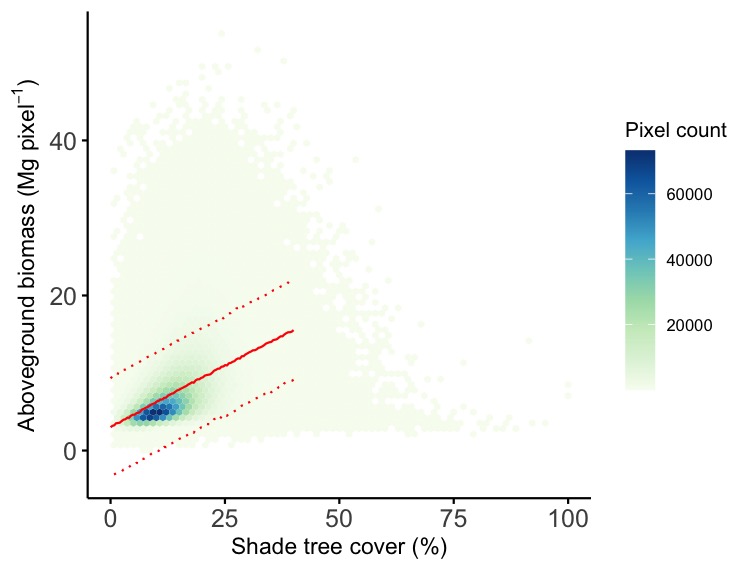}
    \caption{The relationship between shade-tree cover and aboveground biomass (AGB) in cocoa growing areas across Ghana and Côte d’Ivoire.  The plot is based on pixel-level estimates of shade-tree cover and AGB, using a randomly selected 10\% subset of the total dataset (over 3.4 million data points). Data were sampled at a 50-meter ground distance and are presented as a hexbin plot. The hexagonal bins represent the density of data points, with dark blue indicating higher density and light green indicating lower density.  The solid red line shows the posterior median from a Bayesian regression model fit to data with shade cover $\leq$40\%, and the red dotted lines show the 95\% credible interval based on predictions generated from the posterior distribution. Model details are provided in the Supplementary Methods under \textit{Bayesian regression methods for quantifying relationship between cover and biomass}.}
    \label{fig:hexbin_cover_biomass}
\end{figure}

\begin{figure}[ht]
    \centering
    \includegraphics[width=1\textwidth]{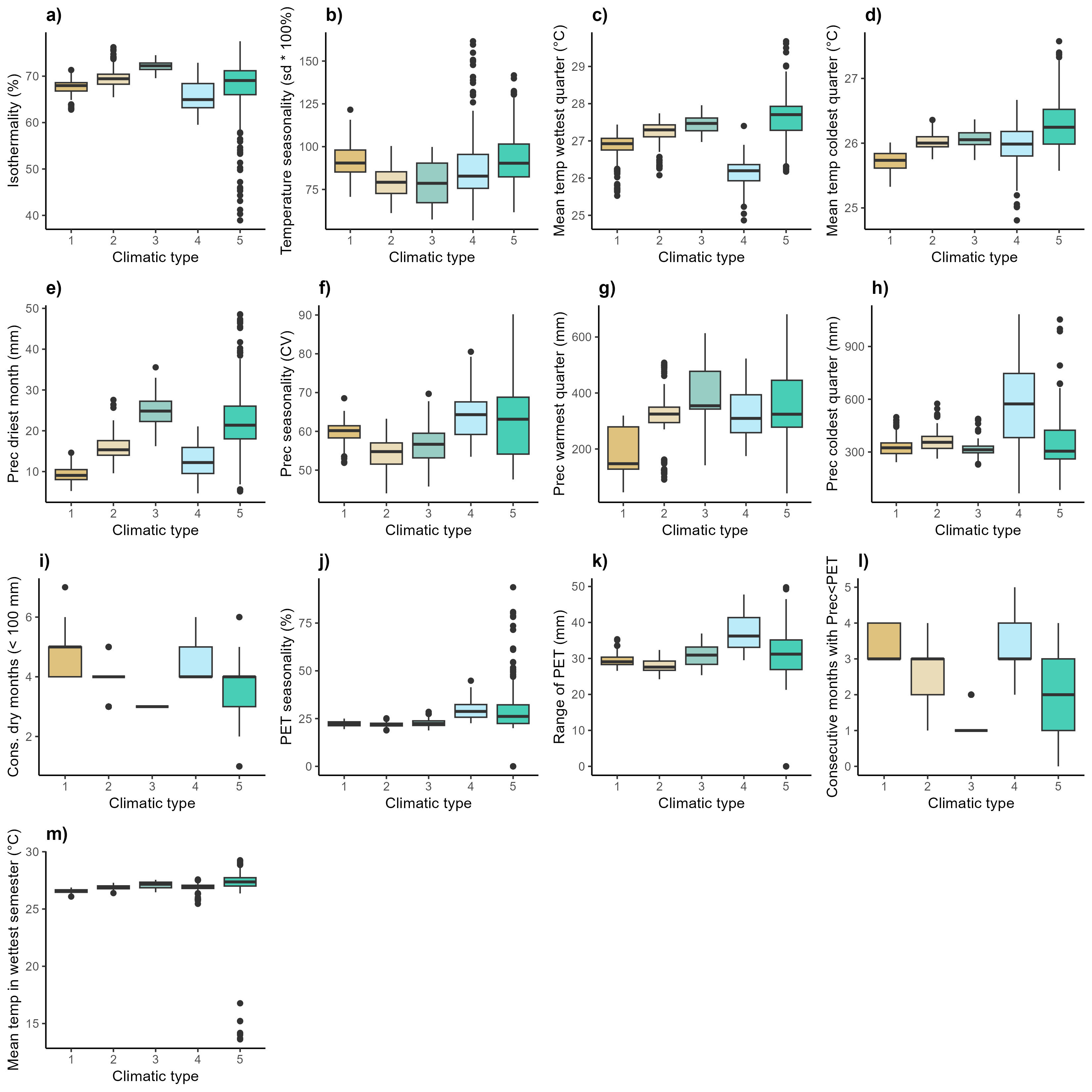}
    \caption{Differences in climatic variable across five distinct agro-climatic types (1–5). The bioclimatic variables shown include: (a) isothermality (BIO 3: ratio of diurnal temperature range to annual temperature range), (b) temperature seasonality (BIO 4: standard deviation of temperature ×100), (c) mean temperature of the wettest quarter (BIO 8), (d) mean temperature of the coldest quarter (BIO 11), (e) precipitation of the driest month (BIO 14), (f) precipitation seasonality (BIO 15: coefficient of variation in monthly precipitation), (g) precipitation of the warmest quarter (BIO 18), (h) precipitation of the coldest quarter (BIO 19), (i) number of consecutive months with precipitation below 100 mm (BIO 20), (j) potential evapotranspiration (PET) seasonality (BIO 26: annual variability in PET), (k) PET range (BIO 29: difference between maximum and minimum PET), (l) number of consecutive months with precipitation less than PET (BIO 21), and (m) mean temperature during the wettest semester (BIO 23). The boxplots display the median (central line), interquartile range (box limits), 10th and 90th percentiles (whiskers), and potential outliers (dots). Different colors correspond to the five agro-climatic types identified.  $n = 936$ cocoa locations across all agro-climatic types. No statistical tests were performed; distributions are shown to visualize climatic differences among clusters identified via Random Forest dissimilarity-based clustering. Bioclimatic variable follow the methods and definitions used in Kamath et al. (2024) \cite{Kamath2024} }
    \label{fig:bioc_boxplot}
\end{figure}

\clearpage

\defaultbibliographystyle{unsrt}
\putbib
\end{bibunit}
\label{LastPagePart2}

\end{document}